\pdfoutput=1

\documentclass[10pt,twocolumn,letterpaper]{article}

\usepackage[pagenumbers]{cvpr}

\usepackage{cuted}    %
\usepackage{currfile} %
\usepackage{caption}  %
\usepackage{amsthm}
\usepackage{multirow}
\usepackage{pifont}
\usepackage{colortbl}
\usepackage{microtype}

\newcommand{\method}{GALILEO\xspace}
\newcommand{\psnrd}{PSNR$_\text{dyn}$\xspace}
\newcommand{\ssimd}{SSIM$_\text{dyn}$\xspace}
\newcommand{\lpipsd}{LPIPS$_\text{dyn}$\xspace}
\newcommand{\T}{^{\top}}
\newcommand{\xh}{\hat{x}}
\DeclareMathOperator{\softmax}{softmax}

\definecolor{dgood}{rgb}{0.10,0.50,0.20}
\definecolor{dbad}{rgb}{0.75,0.15,0.15}
\definecolor{dnoise}{rgb}{0.50,0.50,0.50}
\definecolor{ours}{rgb}{0.114,0.310,0.847}
\newcommand{\up}[1]{{\scriptsize\color{dgood}#1}}
\newcommand{\dn}[1]{{\scriptsize\color{dbad}#1}}
\newcommand{\nn}[1]{{\scriptsize\color{dnoise}#1}}

\newcommand{\best}[1]{\textbf{#1}}

\theoremstyle{plain}
\newtheorem{proposition}{Proposition}

\renewcommand{\paragraph}[1]{\vspace{.35em}\noindent\textbf{#1}}

\AtBeginDocument{\crefname{appendix}{Appendix}{Appendices}\Crefname{appendix}{Appendix}{Appendices}}

\definecolor{cvprblue}{rgb}{0.21,0.49,0.74}
\usepackage[pagebackref,breaklinks,colorlinks,allcolors=cvprblue]{hyperref}

\def\confName{CVPR}
\def\confYear{2027}

\title{GALILEO: Geometric RoPE with Galilean Key Transport for Dynamic Scenes}

\author{
Youngmin Kim$^{*}$ \qquad Seonghyun Jin$^{*}$ \qquad Jong Chul Ye$^{\dagger}$ \\[4pt]
Kim Jaechul Graduate School of Artificial Intelligence, KAIST \\
Seoul, Republic of Korea \\
{\tt\small \{zeromin03,jinotter3,jong.ye\}@kaist.ac.kr} \\[2pt]
{\small $^{*}$Equal contribution \qquad $^{\dagger}$Corresponding author} \\[2pt]
{\small Project page: \url{https://bispl-wm.github.io/GALILEO/}}
}

\begin{document}
\maketitle
\begin{strip}
\centering
\vspace{-2.2em}
\includegraphics[width=\linewidth]{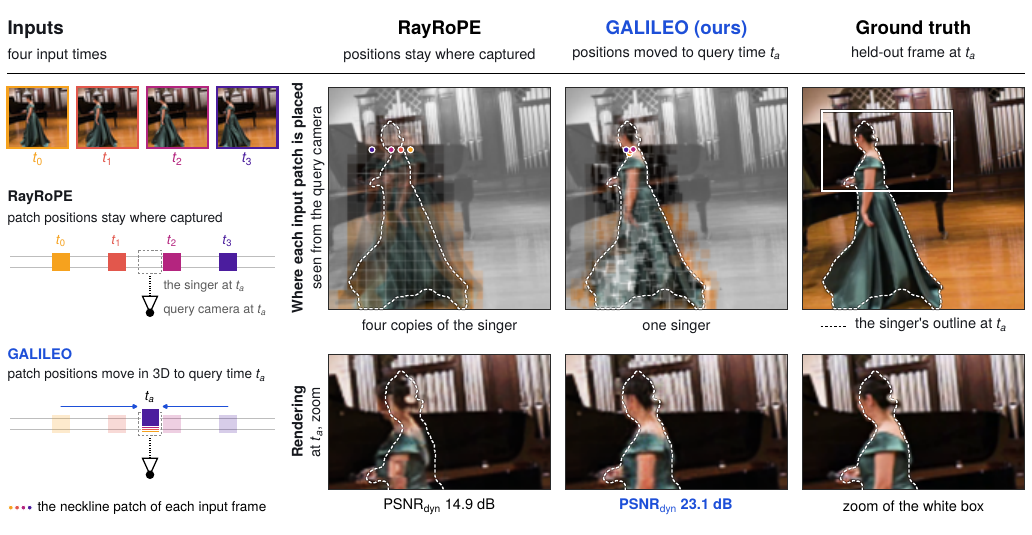}
\captionsetup{hypcap=false}
\captionof{figure}{\textbf{A moving object's patches belong at the query time.}
A novel view synthesis transformer renders the left view at a held-out query time $t_a$.
\emph{Left:} the inputs and a schematic of the idea.
\emph{Top row:} where the encoding places each input patch, at its predicted depth, seen from the query camera (four input times; static patches gray).
RayRoPE~\cite{wu2026rayrope} keeps each patch's 3D position where it was captured; \method moves it to $t_a$ by the patch's estimated 3D velocity.
\emph{Bottom row:} rendering at $t_a$ (\psnrd: PSNR on the moving region).}
\captionsetup{hypcap=true}
\label{fig:teaser}%
\end{strip}

\begin{figure*}[t]
  \centering
  \includegraphics[width=\textwidth]{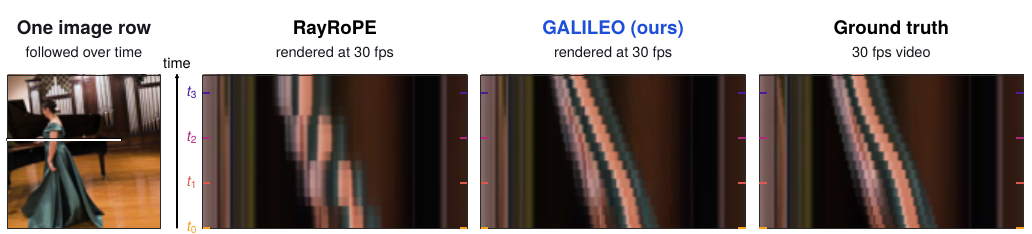}
  \caption{\textbf{A moving object is a line in space--time.} One image row (white line) over time: position runs right, time up, and ticks mark the input times. All are shown at the video's 30\,fps. In the ground truth, the singer traces slanted lines. RayRoPE draws a staircase: its content stays where an input frame captured it, then jumps to the next. \method follows the slanted lines. A second clip is in \cref{sec:supp_qualitative}.}
  \label{fig:xt}
\end{figure*}

\begin{abstract}
Geometry-aware rotary positional encodings, such as RayRoPE, help attention reason about the 3D spatial relationships between tokens and the query camera. However, they implicitly assume a static scene, encoding each key at its observed position regardless of when it was captured. For dynamic scenes with temporally separated input views, this mismatch places the same moving object at a different position in each input frame, so it can be rendered at an inaccurate position. While temporal encodings inform attention \textbf{when} a token was observed, they do not resolve \textbf{where} its content should be at the query time.
We introduce \method, a simple and general approach that aligns positional encodings with dynamic scene geometry through \textbf{Galilean transport}. Before positional encoding, each key's 3D position is transported to the query time using its estimated velocity and the temporal offset. By modifying only key positions, \method integrates directly with existing encodings, including RayRoPE and URoPE, without altering their formulations. Velocities can be obtained from arbitrary sources; by default, we estimate them from token correspondences within the network.
Integrated into LVSM, \method improves dynamic view synthesis on Syn4D, dynamic Objaverse, and real-world Stereo4D videos over baselines. Ablations show that encoding velocities as token features yields smaller gains, while attention visualizations reveal that \method attends to query-time locations of moving content, unlike RayRoPE. Beyond view synthesis, \method also reduces tracking errors on dynamic objects in a pretrained 3D point tracker and improves moving-vehicle generation in a video generator.
\end{abstract}

\section{Introduction}
\label{sec:intro}

Attention is permutation invariant, so a transformer knows where its tokens are only through their positional encoding~\cite{vaswani2017attention}.
Rotary encodings~\cite{su2024roformer,heo2024ropevit} make attention depend on the relative position of two tokens, and a recent line of work makes this relative position three-dimensional.
GTA~\cite{miyato2024gta} and PRoPE~\cite{li2025prope} rotate queries and keys by the relative camera pose, and RayRoPE~\cite{wu2026rayrope} and URoPE~\cite{xie2026urope} also encode where along its camera ray the content of a token lies.
Every position is expressed in the frame of the query camera, so attention depends on how two tokens are related in 3D and not on the choice of world frame.
This matters most where a transformer must reason in 3D without an explicit 3D representation, as in LVSM~\cite{jin2025lvsm}, which renders a target view by attending from target rays to input image patches.

\begin{figure*}[t]
  \centering
  \includegraphics[width=\textwidth]{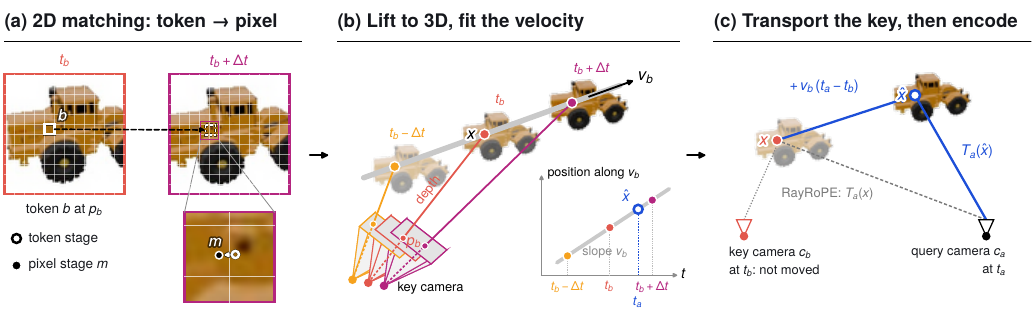}
  \caption{\textbf{Overview of \method.}
  (a)~Each input patch $b$ is matched to its camera's frames at $t_b \pm \Delta t$, from token to pixel accuracy.
  (b)~The patch is lifted to 3D along its ray at the encoding's depth, and so are its matches; a line through them gives a constant velocity $v_b$.
  (c)~For a query at $t_a$, the points of key $b$ are moved to $\xh = x + v_b\,(t_a - t_b)$ then encoded as in RayRoPE. The key camera is not moved.}
  \label{fig:method}
\end{figure*}

These existing geometry-aware encodings, however, implicitly assume that keys and queries correspond to the same instant. Each key is encoded at the 3D position of its content at capture time, even when the query corresponds to a different time. While this distinction is irrelevant for static scenes, it becomes critical for moving objects. Across input frames, the same object occupies different 3D positions, breaking the positional consistency on which RayRoPE relies~\cite{wu2026rayrope} and deviating from the epipolar constraints used by URoPE~\cite{xie2026urope}. As a result, the object may appear at multiple locations in a rendered view (\cref{fig:teaser}), while temporal renderings exhibit stationary copies followed by abrupt jumps (\cref{fig:xt}). In short, existing encodings tell attention where moving content \emph{was}, rather than where it \emph{is}.

A common approach is to incorporate temporal information through temporal RoPE~\cite{wang2026bullettime} or time embeddings~\cite{liang2025btimer,wu2025cat4d}. However, such signals indicate when a key was observed, not where its content is at the query time. Since all keys from the same frame share the same temporal information regardless of their motion, these approaches do not resolve the underlying geometric mismatch. Indeed, all our baselines incorporate temporal information but still encode moving keys at their observed positions.

In contrast, we argue that a geometry-aware encoding should place each key where its content is at the query time. 
To achieve this, we introduce \method, a novel method that incorporates Galilean transport into positional encoding.
 For a key $b$ observed at time $t_b$ and a query at time $t_a$, its 3D points are shifted by $v_b(t_a-t_b)$, where $v_b$ denotes the velocity of its content (\cref{fig:method}).
 This transport provides a first-order approximation of motion over the short intervals between input frames. 
 By modifying only key positions, \method leaves queries and the attention operator unchanged, preserves the original encoding for stationary keys, and integrates directly with RayRoPE and URoPE. The method supports velocities from any source; by default, \method estimates them within the network by matching tokens across adjacent frames and lifting the correspondences to 3D using the depths employed by the encoding.

Integrated into LVSM, \method improves the rendering of moving content over baselines on synthetic Syn4D~\cite{jiang2026syn4d} and dynamic Objaverse~\cite{deitke2023objaverse} scenes, as well as real-world Stereo4D~\cite{jin2025stereo4d} videos.
It also carries over to other models: attached to the pretrained tracker ARROW~\cite{fradlin2026arrow}, it lowers 3D tracking error on dynamic objects, and in a 5B-parameter video generator that renders driving scenes from their geometry~\cite{mirros2026agentgarten,wan2025wan}, it improves moving objects with velocities from their 3D boxes.

Further analyses suggest that these improvements arise from aligning key positions with moving content, rather than simply providing motion information as additional features. Attention visualizations show that \method attends more to the actual locations of moving content, unlike RayRoPE, which favors positions implied by static geometry (\cref{fig:attention}). Moreover, \method is far more robust to changes in the reference frame, even when the underlying scene remains unchanged (\cref{fig:boost}).

Our contributions are summarized as follows:
\begin{itemize}[leftmargin=1.2em,itemsep=0pt,topsep=2pt]
    \item We identify a key limitation of existing geometry-aware rotary encodings in dynamic scenes: they encode moving content at its observed position, not its query-time position, which temporal embeddings alone cannot resolve.
    \item We introduce \method, a Galilean transport of keys inside the encoding that places each key where its content is at the query time. It integrates directly with RayRoPE and URoPE without modifying queries or the attention operator, and supports motion from arbitrary sources.
    \item We demonstrate improvements in dynamic view synthesis across three datasets, as well as in 3D tracking and video generation. Further analyses highlight the role of moving keys to the query-time position through attention visualizations and robustness tests under Galilean boosts.
\end{itemize}

\section{Related Work}
\label{sec:related}

\paragraph{Geometry-aware positional encodings.}
Feed-forward view synthesis models~\cite{sajjadi2022srt,hong2024lrm,jin2025lvsm} leave much of the geometry to how tokens are positioned.
From absolute rays~\cite{sitzmann2021lfn} and 3D positions~\cite{liu2022petr,xiong2023cape}, encoding methods have evolved to relative ones built on rotary embeddings~\cite{su2024roformer,heo2024ropevit}. For example, GTA~\cite{miyato2024gta} and PRoPE~\cite{li2025prope} transform queries, keys, and values by the relative camera pose; RayRoPE~\cite{wu2026rayrope} and URoPE~\cite{xie2026urope} also encode where along its ray a token's content lies. Modern video generators adopt such encodings for camera control~\cite{jin2026crepe,li2026rerope,qiao2026merope} and world models~\cite{xiang2026georope}.

Relative encodings make attention invariant to rigid transformations of the world coordinate frame, unlike absolute encodings. %
To account for time,
temporal RoPE~\cite{wang2024qwen2vl,wang2026bullettime} and time embeddings~\cite{liang2025btimer,wu2025cat4d} encode the observation time axis. However, they do not account for content motion, as all tokens within a frame share the same temporal information. In contrast, \method transports each key to its query-time position, ensuring that, given accurate motion estimates, its content is encoded at the correct position relative to the query camera, regardless of the reference frame, including moving ones (\cref{prop:boost}).

\paragraph{Warping positional encodings.}
Training-free motion control in video diffusion transformers often warps rotary embeddings or queries along a desired 2D motion~\cite{gokmen2025ropecraft,lei2025ditraj,choo2026qwerty}.
PE-Field~\cite{bai2026pefield,bai2026pefield4d} and UCM~\cite{xu2026ucm} instead reproject the codes of context tokens into the target view with estimated depth and camera poses.
However, these reprojections account for camera motion rather than the motion of the content itself: PE-Field 4D reprojects each source frame only to a target view at the same time, while UCM trains its model to disregard moving objects.
In contrast, \method transports each key according to its content's 3D motion to the query time, while encoding its position relative to the query camera.
Whereas camera-only reprojection leaves moving content at its capture-time positions (\cref{fig:teaser}),
\method aligns its observations at a common query-time position (\cref{prop:coobs}), allowing attention to aggregate them (\cref{fig:attention}).

\paragraph{Motion in dynamic view synthesis.}
4D scene representations model motion within the representation itself, through deformation or scene-flow fields~\cite{li2021nsff,park2021nerfies,pumarola2021dnerf}, space--time planes~\cite{cao2023hexplane,fridovichkeil2023kplanes}, or moving Gaussians~\cite{luiten2024dynamic3dgs,wu20244dgs}, optimized per scene or predicted feed-forward~\cite{balice2026nopose4d,lin2026movies,ma20254dlrm,xu20254dgt}; other feed-forward models regress a separate set of Gaussians for each target time~\cite{liang2025btimer,ren2024l4gm}.
In contrast, renderers that predict target views directly~\cite{elmieh2026online,ma2026fastmemory,tanay2026grvs} lack an explicit representation to deform, while WildRayZer~\cite{chen2026wildrayzer} handles dynamic content by masking it out.
Motion compensation offers another approach, warping features or sampling locations, as in video codecs~\cite{wiegand2003h264}, flow-aligned restoration~\cite{chan2022basicvsrpp}, deformable and trajectory attention~\cite{zhu2021deformabledetr,patrick2021trajectory}, and DynIBaR~\cite{li2023dynibar}.
Unlike these approaches, \method incorporates scene flow into positional encoding, without requiring an explicit scene representation or resampling content. This allows seamless integration with existing token-based models (\cref{sec:transfer}), avoids warping holes, and supports arbitrary query times using the same keys (\cref{fig:xt}).

\section{\method: Galilean Key Transport}
\label{sec:method}

\subsection{Geometry-aware rotary encodings}
\label{sec:prelim}

Consider a  model that receives input views with known cameras, captured at times $t_n$, and renders a target view for a given camera and time.
Every view is split into patch tokens, and the transformer attends over the tokens of all views.
Each token inherits the camera and timestamp of its view. Specifically, query $a$ is associated with camera center $c_a$, rotation $R_a$, and time $t_a$, and key $b$ is defined analogously.

We build on ray-based encodings~\cite{wu2026rayrope,xie2026urope}, which describe each token by its camera center $c_b$ and a set of 3D points $X_b$ along its rays, where its content may lie.
Specifically, for query $a$, the points are expressed in its camera frame:
\begin{equation}
  T_a(x) = R_a\T (x - c_a),
  \label{eq:rel}
\end{equation}
and mapped to cosine and sine coefficients at a set of frequencies.
The coefficients form a block-diagonal matrix $M_a(b)$ of $2\times2$ rotations, and attention between query $a$ and key $b$ uses the logit
\begin{equation}
  \ell_{ab} = \big(M_a(a)\T q_a\big)\T \big(M_a(b)\T k_b\big) / \sqrt{d},
  \label{eq:logit}
\end{equation}
with head dimension $d$.
Values are transformed by the same matrix as keys, and the output is rotated back by $M_a(a)$.
Accordingly, expressing coordinates relative to the query camera makes attention invariant to rigid transformations of the world coordinate frame. 

The base encodings used in this paper also incorporate temporal RoPE, with selected rotation blocks applying phases proportional to $t_a-t_b$  (\cref{sec:supp_details}).
Each input or target view serves as a query view in turn. Since its tokens share the same camera and timestamp, the key encoding matrices can be computed once per query view  (\cref{sec:supp_method}).

\begin{table*}[t]
  \centering
  \scriptsize
  \setlength{\tabcolsep}{3pt}
  \resizebox{\textwidth}{!}{%
  \begin{tabular}{@{}l ccccc @{\hspace{8pt}} ccccc @{\hspace{8pt}} ccccc@{}}
    \toprule
    & \multicolumn{5}{c}{Stereo4D (real)} & \multicolumn{5}{c}{Objaverse} & \multicolumn{5}{c}{Syn4D} \\
    \cmidrule(lr){2-6} \cmidrule(lr){7-11} \cmidrule(l){12-16}
    & \multicolumn{2}{c}{image} & \multicolumn{3}{c}{video} & \multicolumn{2}{c}{image} & \multicolumn{3}{c}{video} & \multicolumn{2}{c}{image} & \multicolumn{3}{c}{video} \\
    \cmidrule(lr){2-3} \cmidrule(lr){4-6} \cmidrule(lr){7-8} \cmidrule(lr){9-11} \cmidrule(lr){12-13} \cmidrule(l){14-16}
    Method & PSNR$_\mathrm{dyn}\!\uparrow$ & PSNR$\uparrow$ & tOF$\downarrow$ & $\delta_\mathrm{avg}\!\uparrow$ & JEDi$\downarrow$ & PSNR$_\mathrm{dyn}\!\uparrow$ & PSNR$\uparrow$ & tOF$\downarrow$ & $\delta_\mathrm{avg}\!\uparrow$ & JEDi$\downarrow$ & PSNR$_\mathrm{dyn}\!\uparrow$ & PSNR$\uparrow$ & tOF$\downarrow$ & $\delta_\mathrm{avg}\!\uparrow$ & JEDi$\downarrow$ \\
    \midrule
    Plücker & 23.59 & 29.17 & 1.52 & .918 & 1.047 & -- & -- & -- & -- & -- & 19.20 & 21.72 & 2.79 & .616 & 2.632 \\
    PRoPE & 24.06 & 29.95 & 1.44 & .926 & 0.741 & 16.73 & 27.51 & 1.31 & .748 & 0.250 & 19.34 & 22.57 & 2.39 & .635 & 2.590 \\
    GTA & 24.15 & 30.07 & 1.41 & .928 & 0.726 & 16.81 & 27.76 & 1.41 & .755 & 0.184 & 19.28 & 22.46 & 2.31 & .632 & 2.789 \\
    \midrule
    URoPE & 24.61 & 30.73 & 1.08 & .939 & 0.493 & 16.33 & 27.07 & 1.44 & .730 & 0.236 & 19.66 & 22.99 & 2.18 & .627 & 1.410 \\
    \quad + GALILEO & 25.41 & 31.00 & 0.89 & .950 & 0.391 & 16.70 & 27.25 & 1.30 & .739 & 0.197 & 20.14 & 23.11 & 1.82 & .637 & 0.787 \\
    \quad $\Delta$ & \,\up{+0.79} & \,\up{+0.27} & \,\up{$-$0.19} & \,\up{+.011} & \,\up{$-$0.102} & \,\up{+0.37} & \,\up{+0.18} & \,\up{$-$0.14} & \,\up{+.010} & \,\up{$-$0.039} & \,\up{+0.49} & \,\up{+0.12} & \,\up{$-$0.37} & \,\up{+.010} & \,\up{$-$0.622} \\
    \midrule
    RayRoPE & 24.61 & 31.04 & 1.04 & .947 & 0.466 & 16.67 & 27.42 & 1.53 & .743 & 0.346 & 19.88 & 23.17 & 2.19 & .639 & 1.075 \\
    \quad + GALILEO & 25.78 & 31.34 & 0.80 & .958 & 0.343 & 17.96 & 28.25 & 1.12 & .770 & 0.268 & 20.82 & 23.51 & 1.65 & .668 & 0.380 \\
    \quad $\Delta$ & \,\up{+1.16} & \,\up{+0.30} & \,\up{$-$0.24} & \,\up{+.010} & \,\up{$-$0.122} & \,\up{+1.29} & \,\up{+0.83} & \,\up{$-$0.41} & \,\up{+.026} & \,\up{$-$0.078} & \,\up{+0.94} & \,\up{+0.34} & \,\up{$-$0.53} & \,\up{+.029} & \,\up{$-$0.695} \\
    \bottomrule
  \end{tabular}}
  \caption{\textbf{Main results} on three datasets (metrics and averaging in \cref{sec:setup}).
  Every method carries a time signal, a temporal RoPE or, for Plücker, a time embedding, and each + \method model differs from its base encoding only in the motion components.
  $\Delta$: change from the base encoding; every change is beyond noise (90\% bootstrap interval, or a paired permutation test for JEDi; \cref{sec:supp_details}).
  --: Plücker failed to train on Objaverse.
  All image metrics are in \cref{tab:full}.}
  \label{tab:main}
\end{table*}

\subsection{Key transport}
\label{sec:transport}

\paragraph{The issue.}
Consider moving content observed at different times. The points in $X_b$ represent the positions of key $b$'s content at its observation time $t_b$, whereas \cref{eq:rel} expresses these positions relative to the query camera at time $t_a$. This yields the correct query-relative position only if the content remains stationary between the two instants. For content following a trajectory $x(t)$, however, the query requires its position at time $t_a$, expressed as $T_a(x(t_a))$. Thus, while \cref{eq:rel} is sufficient for static content, it encodes outdated positions for moving content.

\paragraph{Galilean transport.}
We approximate the motion of content over the interval between a key and a query using a first-order model.
Let $v_x$ be the 3D velocity of the content at point $x \in X_b$, in world coordinates.
Before encoding key $b$ for a query at time $t_a$, we move its points to that time,
\begin{equation}
  \xh = x + v_x\,(t_a - t_b), \qquad x \in X_b,
  \label{eq:transport}
\end{equation}
and compute the key's coefficients and its matrix $\hat{M}_a(b)$, which transforms the key and its value, from $T_a(\xh)$ instead of $T_a(x)$.
In RayRoPE, all points of a token share its velocity, whereas in URoPE, each depth anchor has its own velocity.

The camera center $c_b$ remains unchanged, as it represents the observer's position rather than that of the content. Target tokens, which carry no observed content, are not transported, and query encodings remain unchanged. Since key encodings are already computed per query view, the transport requires only shifting key points before encoding, allowing the same key to be encoded at different positions for different query times. For static content ($v_x=0$) or when $t_a=t_b$, the method reduces exactly to the original encoding, with temporal dependence determined solely by $t_a-t_b$.

\subsection{Theoretical findings}

RayRoPE is designed to assign similar positional encodings to patches observing the same 3D point~\cite{wu2026rayrope}. Galilean transport restores this property for moving content.
The four dots on the neckline in \cref{fig:teaser} illustrate such observations.
\begin{proposition}[Co-observation]
Suppose keys $b$ and $b'$ observe the same 3D point moving at constant velocity $v$. With this velocity used for transport, their content points are both mapped to $T_a\big(x(t_a)\big)$ for any query $a$.
  \label{prop:coobs}
\end{proposition}
\begin{proof}
This follows from $x(t_b) + v\,(t_a-t_b) = x(t_a)$, whereas without transport, the query-relative positions of the two observations differ by $R_a\T v\,(t_b-t_{b'})$. 
\end{proof}

The relative transformation in \cref{eq:rel} makes the encoding invariant to rigid changes in the origin and orientation of the world frame. With temporally separated observations, however, the reference frame may itself move, as when camera poses are expressed relative to a moving vehicle.
\begin{proposition}[Galilean boost]
Consider a change of reference frame that shifts all camera and content positions at time $t$ by $w\,t$. 
  Then the transported coordinates $T_a(\xh)$ are unchanged, and the static coordinates $T_a(x)$ change by $-R_a\T w\,(t_a - t_b)$.
  \label{prop:boost}
\end{proposition}

\noindent Under this transformation, velocities estimated from 3D positions become $v_x+w$, yielding 
\[
  (x + w t_b) + (v_x + w)(t_a - t_b) - (c_a + w t_a) = \xh - c_a .
\]
With true content displacements, the same principle extends to arbitrary time-dependent rigid changes of reference frame. For Galilean boosts, even least-squares velocity estimates preserve this invariance regardless of the underlying content motion. Thus, while static encoding correctly aligns only content stationary in the given reference frame, transported content coordinates are invariant to Galilean boosts. This property applies to content points, not to the key camera center, which remains untransported and changes by $R_a\T w\,(t_b-t_a)$. We evaluate the resulting robustness of trained models in \cref{sec:boost}.

Our theoretical understanding clarifies the difference from temporal RoPE.
Temporal RoPE encodes when a key was observed, assigning the same temporal information to all keys from a frame regardless of their motion. 
Specifically, the pair $\big(T_a(\xh),\, t_a - t_b\big)$ describes the content relative to the query in both space and time.
Temporal RoPE provides the time offset $t_a-t_b$, which \method preserves, but cannot supply the first ($T_a(\xh)$).
In contrast, Galilean transport adjusts where each key's content is encoded. 

\subsection{Velocity estimation}
\label{sec:velocity}

While \method supports velocities from arbitrary sources, including optical flow and 3D tracks (\cref{sec:transfer}), we estimate them within the network by default, without relying on external models at test time (\cref{fig:method}a,b).
Specifically, each input token is matched to adjacent frames from the same camera using a coarse-to-fine approach~\cite{sun2018pwcnet,sun2021loftr}: initial correspondences are established at token resolution using transformer features and refined to pixel accuracy with a lightweight CNN.
The matched points are then lifted to 3D using camera poses and the encoding's depths, accounting for camera motion, and their velocities are fit by least squares under a constant-velocity model (\cref{sec:supp_velocity}).

\paragraph{Depth errors.}
Depth errors can be amplified when estimating velocity from closely spaced frames.
A depth error $\delta$ at one end of a correspondence introduces a transported-position error of approximately $\delta\,|t_a-t_b|/\Delta t$, where $\Delta t$ is the interval between adjacent frames. 
This error lies primarily along the capturing camera's viewing ray, making it less apparent from that viewpoint but potentially significant from others. 
Thus, accurate depth is important for reliable transport (\cref{sec:where}), which motivates depth supervision.

\subsection{Training}
\label{sec:training}

We train \method with the backbone's image loss, a matching loss, and a depth loss:
\begin{equation}
  \mathcal{L} = \mathcal{L}_\text{img} + \mathcal{L}_\text{match} + \mathcal{L}_\text{depth},
  \label{eq:loss}
\end{equation}
where $\mathcal{L}_\text{match}$ supervises both stages of matching using ground-truth correspondences, and $\mathcal{L}_\text{depth}$ supervises the depths used in positional encoding to mitigate errors amplified during velocity estimation.
Although the original RayRoPE does not use depth supervision, we apply the same depth losses to its baseline in our experiments to isolate the effect of motion transport. Loss definitions and weights are provided in \cref{sec:supp_method}.

To mitigate instability in velocity predictions early in training, we use ground-truth velocities for key transport during the first $s_0$ steps. We then gradually transition to predicted velocities through linear interpolation, $v = (1-\alpha)v_\text{gt} + \alpha v_\text{pred}$, where $\alpha$ increases from 0 at step $s_0$ to 1 at step $s_1$. Thereafter, and at test time, only predicted velocities are used. 
This schedule is inspired by mix-forcing~\cite{ye2026yonosplat,jin2026crepe} and related to scheduled sampling~\cite{bengio2015scheduled}.

\section{Experiments}
\label{sec:experiments}

\subsection{Setup}
\label{sec:setup}

\paragraph{Data.}
We evaluate \method on three datasets spanning synthetic and real-world data, from individual objects to complex scenes: dynamic Objaverse~\cite{deitke2023objaverse}, consisting of animated objects rerendered by us; Syn4D~\cite{jiang2026syn4d}, featuring synthetic scenes with a person and multiple animated objects; and Stereo4D~\cite{jin2025stereo4d}, containing real-world stereo videos.
Inputs are a few frames from two cameras over a short interval, and targets are other frames of the same test clip: novel times from the input cameras and, except on Stereo4D, which has only two cameras, novel views from new cameras. Test clips are disjoint from training (\cref{sec:supp_details}).

\paragraph{Metrics.}
We evaluate image quality using PSNR on dynamic regions (\psnrd), defined as pixels displaced by more than 1.5\,px relative to the nearest input time, and full-image PSNR, with SSIM and LPIPS in the supplement. For videos rendered over each clip, we assess temporal quality using tOF~\cite{chu2020tecogan}, which measures agreement with ground-truth motion; $\delta_\mathrm{avg}$~\cite{doersch2022tapvid}, which evaluates how accurately a tracker~\cite{karaev2024cotracker3} follows moving points; and JEDi~\cite{luo2025jedi}, which compares distributions of V-JEPA~\cite{bardes2024vjepa} features sensitive to temporal distortions. Results average novel views and novel times with equal weight, and changes count as significant when their 90\% confidence intervals exclude zero (\cref{sec:supp_details}).

\paragraph{Baselines and training.}
All methods share the LVSM backbone and training recipe, are trained from scratch for 20k steps, and differ only in the positional encoding: Pl\"ucker ray tokens with a time embedding, and PRoPE~\cite{li2025prope}, GTA~\cite{miyato2024gta}, URoPE~\cite{xie2026urope}, and RayRoPE~\cite{wu2026rayrope}, each with a temporal RoPE.
Unless stated otherwise, \method is applied to RayRoPE, and RayRoPE receives the same depth losses.
No method uses ground-truth depth, flow, or motion masks at test time (details in \cref{sec:supp_details}).

\subsection{Main results}
\label{sec:main}
\begin{figure*}[t]
  \centering
  \includegraphics[width=\textwidth]{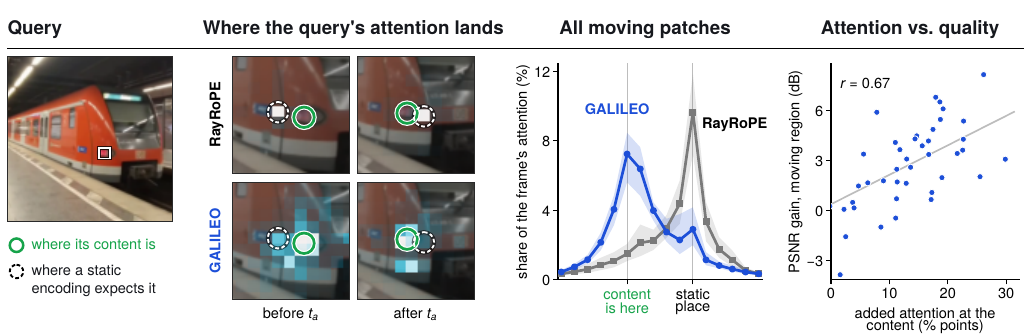}
  \caption{\textbf{Attention follows the encoding.} \emph{Left:} a query patch on the moving train at the held-out time $t_a$ and its attention in the two input frames farthest in time. Green circle: where the patch's content is in that frame, found by optical flow from the ground-truth frame at $t_a$. Dashed circle: where a static scene would put it, the patch's 3D point at $t_a$ projected into the frame. \emph{Middle:} attention of all moving patches along the line through the two places. \emph{Right:} per held-out frame, the attention that \method adds at the content predicts its gain in \psnrd.}
  \label{fig:attention}
\end{figure*}

\cref{tab:main} summarizes the results with \cref{tab:full} for the remaining image metrics.
On both RayRoPE and URoPE, \method improves every metric of \cref{tab:main} on all three datasets, including the real videos of Stereo4D, and raises \psnrd by 0.9--1.3\,dB on RayRoPE and 0.4--0.8\,dB on URoPE.
The video metrics indicate that the gain is not only in sharpness but also in motion: rendered content moves more like the real content (tOF), a tracker follows its points more closely ($\delta_\mathrm{avg}$), and JEDi falls by about two thirds with RayRoPE on Syn4D.
The gains persist with four times longer training on Syn4D (\cref{tab:longer}): at 80k steps, \method still improves every metric, and RayRoPE at 80k stays below \method at 20k in \psnrd and in every video metric.
They also persist with twice as many layers on Stereo4D (\cref{tab:deeper}).
        
Both models render any time in the input window; over that window, RayRoPE renders a moving object as a staircase and \method as the slanted line of the ground truth (\cref{fig:xt}).

\subsection{Ablation studies}
\label{sec:where}

\begin{table}[t]
  \centering
  \footnotesize
  \setlength{\tabcolsep}{2.5pt}
  \resizebox{\columnwidth}{!}{%
  \begin{tabular}{@{}l cc cc@{}}
    \toprule
    & \multicolumn{2}{c}{image} & \multicolumn{2}{c}{video} \\
    \cmidrule(lr){2-3} \cmidrule(l){4-5}
    Model & PSNR$_\mathrm{dyn}\!\uparrow$ & PSNR$\uparrow$ & $\delta_\mathrm{avg}\!\uparrow$ & JEDi$\downarrow$ \\
    \midrule
    \multicolumn{5}{@{}l}{\textit{Syn4D} ($\Delta$ vs.\ (a); predicted warp vs.\ (b))} \\
    RayRoPE (a) & 19.88 & 23.17 & .639 & 1.075 \\
    RayRoPE, matching losses & 20.11\,\up{+0.23} & 23.38\,\up{+0.21} & .643\,\up{+.004} & 1.008\,\up{$-$0.067} \\
    RayRoPE, velocity features & 20.10\,\up{+0.22} & 23.38\,\up{+0.21} & .644\,\up{+.005} & 0.993\,\up{$-$0.082} \\
    GALILEO (b) & 20.82\,\up{+0.94} & 23.51\,\up{+0.34} & .668\,\up{+.029} & 0.380\,\up{$-$0.695} \\
    GALILEO, GT motion & 21.26\,\up{+1.38} & 23.66\,\up{+0.49} & .694\,\up{+.055} & 0.398\,\up{$-$0.677} \\
    RayRoPE, predicted warp & 20.58\,\dn{$-$0.24} & 23.26\,\dn{$-$0.25} & .669\,\nn{+.001} & 0.673\,\dn{+0.293} \\
    \midrule
    \multicolumn{5}{@{}l}{\textit{Syn4D, GT depth} ($\Delta$ vs.\ (a); GT warp vs.\ (c))} \\
    RayRoPE & 21.43\,\up{+1.55} & 25.09\,\up{+1.92} & .705\,\up{+.066} & 0.603\,\up{$-$0.472} \\
    GALILEO, GT motion (c) & 23.75\,\up{+3.87} & 25.66\,\up{+2.49} & .786\,\up{+.147} & 0.131\,\up{$-$0.944} \\
    RayRoPE, GT warp & 23.85\,\nn{+0.11} & 25.30\,\dn{$-$0.36} & .775\,\dn{$-$.011} & 0.273\,\dn{+0.142} \\
    \midrule
    \multicolumn{5}{@{}l}{\textit{Motion off at test} ($\Delta$ vs.\ the model with motion)} \\
    Syn4D & 19.84\,\dn{$-$0.98} & 23.29\,\dn{$-$0.22} & .641\,\dn{$-$.027} & 1.159\,\dn{+0.779} \\
    Syn4D, GT depth and motion & 21.06\,\dn{$-$2.69} & 24.91\,\dn{$-$0.75} & .681\,\dn{$-$.105} & 0.575\,\dn{+0.445} \\
    Objaverse & 16.15\,\dn{$-$1.81} & 27.61\,\dn{$-$0.64} & .745\,\dn{$-$.025} & 0.393\,\dn{+0.125} \\
    Stereo4D & 24.47\,\dn{$-$1.31} & 30.95\,\dn{$-$0.39} & .947\,\dn{$-$.011} & 0.514\,\dn{+0.171} \\
    \bottomrule
  \end{tabular}}
  \caption{\textbf{Motion and depth} (metrics as in \cref{tab:main}; rows explained in \cref{sec:where}).
  Small numbers: change from the reference named in each block (green: better, red: worse, gray: within noise).
  GT: ground truth, in training and test.
  GT / predicted warp: inputs warped to the query time instead.
  Other image metrics are in \cref{tab:ablations}.}
  \label{tab:motion}
\end{table}

\cref{tab:motion} disentangles the contributions of key transport, motion information, supervision, and temporal encoding.
First,  to determine whether the gains arise from motion information or key placement, we compare \method with two alternatives: training the matcher without transporting keys, and providing the estimated velocities as additional token features. Both yield only marginal improvements in \psnrd and little change in motion metrics, whereas transporting keys with the same velocities improves all metrics substantially. Unlike token features, which remain fixed across queries, the transport adjusts each key's position to the query time.
Second, even with image reconstruction loss alone, the transport improves \psnrd by 0.56\,dB, accounting for 60\% of \method's gain. Additional supervision contributes the remaining 40\% and improves LPIPS (\cref{sec:supp_ablations}). Token-level matching already improves \psnrd, with the largest additional gain coming from restricting matches to adjacent frames (\cref{sec:supp_velocity}).
Disabling transport at test time reduces \psnrd and all video metrics to RayRoPE's level or worse across all three datasets. This confirms that the trained model relies on the transported key positions rather than merely benefiting from motion-related supervision.

The gain of \method grows as keys are placed more accurately. We train and evaluate \method using ground-truth motion to transport keys, with predicted or ground-truth depth. Ground-truth motion further improves \psnrd, while the estimated velocities recover about two-thirds of its gain over RayRoPE. Depth also limits transport: even with correct motion, inaccurate depth misplaces transported keys. With ground-truth depth, ground-truth motion improves \psnrd by 2.3\,dB, compared with 1.38\,dB using predicted depth (\cref{tab:motion,fig:headroom}).

We further compare \method with two variants of warping the input images to the query time, using ground-truth depth and motion or those predicted by \method. The ground-truth comparison favors warping: with the same per-token motion and a ground-truth mask of the moving pixels, it moves only those pixels within each token, whereas our transport moves each key as a whole. Still, it only matches \method in \psnrd and is worse in full-image PSNR and all motion metrics, with twice the JEDi. The predicted variant falls below \method in both PSNRs and JEDi (\cref{tab:motion}).

\section{Discussion}

\paragraph{Attention follows the encoding.}
\label{sec:attention}
To examine whether the gains arise from key placement, we analyze where target tokens attend in the input views (\cref{fig:attention}). We compare attention at two locations: the \emph{content place}, where the target content appears in an input frame, estimated using optical flow~\cite{teed2020raft} from the ground-truth target frame; and the \emph{static place}, where the content would appear under a static-scene assumption, obtained by projecting its 3D point at the ground-truth depth into the input frame.

The two places coincide for static content, to 0.2\,px in the median, and separate for moving content. On 80 Stereo4D test clips, we take every pair of a moving target token and an input view whose two places are at least 10\,px apart, and measure the share of the frame's attention within a disc around each place, of radius half their distance (\cref{sec:supp_attention}).

RayRoPE puts 20.8\% of a frame's attention at the static place and 6.9\% at the content; \method reverses this, with 22.2\% at the content and 8.8\% at the static place.
On static content, the two models attend alike.
The first layer of \method, which runs before the velocities exist, still attends like RayRoPE; the shift appears in the later layers (\cref{fig:layers}).
Across held-out frames, the added attention at the content place correlates with the gain in \psnrd.

\begin{figure}[t]
  \centering
  \includegraphics[width=\linewidth]{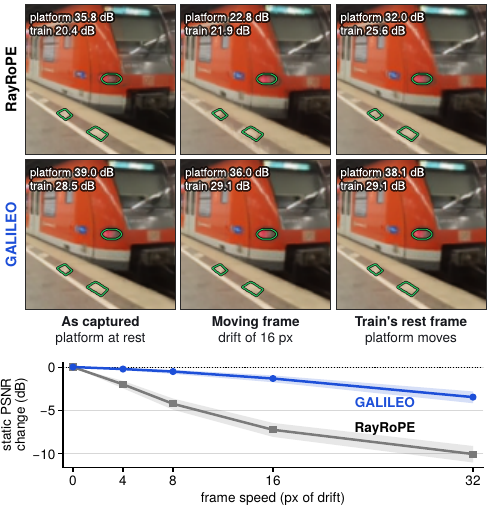}
  \caption{\textbf{The Galilean boost test.} The same clip described from a frame moving at constant velocity $w$: every camera position is shifted by $w\,t$, and nothing else changes. \emph{Top:} RayRoPE renders well only what is at rest in the frame of its poses; \method renders the platform and the train in every frame. Green: the lamp and two platform tiles in the ground truth. \emph{Bottom:} change in static-region PSNR with frame speed.}
  \label{fig:boost}
\end{figure}

\paragraph{Robustness to Galilean boosts.}
\label{sec:boost}
We test \cref{prop:boost} by expressing the same clips in a moving reference frame, without requiring additional data. Specifically, we shift all input and target camera centers by $w\,t$, while keeping images, rotations, intrinsics, and timestamps unchanged. Since the physical scene and ground-truth images remain identical, and camera poses enter the models only through positional encoding, any change in the output can be attributed to the encoding. We choose $w$ along the image $x$-axis of the first camera and express its magnitude as the resulting drift at the median scene depth over the input window.
\cref{fig:boost} (bottom) shows the static-region PSNR as a function of frame speed.
In a moving frame, static content has a nonzero velocity, which RayRoPE does not account for; at a drift of 16\,px, its static-region PSNR drops by 7.2\,dB.
\method, whose estimated velocities include the frame motion, loses 1.3\,dB.
In the clip of \cref{fig:boost} (top), setting $w$ to minus the velocity of the train brings the train to rest and sets the platform in motion.
RayRoPE then gains 5.2\,dB on the train and loses 3.8\,dB on the platform, while \method changes by less than 1\,dB in both regions.

Although far more robust, \method is not exactly invariant: its first layer runs without velocities, and key camera centers are not transported (\cref{prop:boost}).

\begin{table}[t]
  \centering
  \footnotesize
  \setlength{\tabcolsep}{2pt}
  \begin{tabular}{l ccc c ccc}
    \toprule
    & \multicolumn{3}{c}{EPE$\downarrow$} && \multicolumn{3}{c}{APD$\uparrow$} \\
    \cmidrule(lr){2-4} \cmidrule(lr){6-8}
    & Input & Between & After && Input & Between & After \\
    \midrule
    ARROW & 4.89 & 6.86 & 8.67 && 44.6 & 35.6 & 29.2 \\
    + plain adapter & 4.29 & 6.08 & 7.64 && 47.8 & 38.5 & 32.0 \\
    + RayRoPE & 3.82 & 4.99 & 7.16 && 50.9 & 43.5 & 33.8 \\
    + RayRoPE + GALILEO & \textbf{3.66} & \textbf{4.47} & \textbf{6.68} && \textbf{51.8} & \textbf{46.2} & \textbf{35.6} \\
    \bottomrule
  \end{tabular}
  \caption{\textbf{3D tracking with ARROW} on dynamic Objaverse; each adapter is added to frozen ARROW.
  At the input times, every query point is scored; between and after them, points that move more than 1.5\,px.
  EPE is the 3D end-point error and APD the mean share of points within 1--16\%, both relative to object size (\cref{sec:supp_tracking}).}
  \label{tab:tracking}
\end{table}

\begin{figure}[t]
  \centering
  \includegraphics[width=\linewidth]{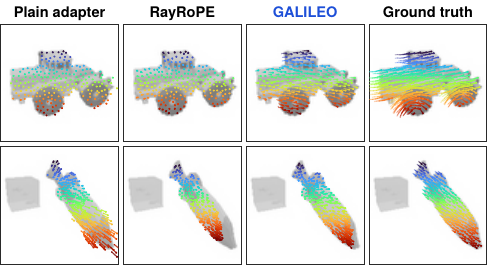}
  \caption{\textbf{3D tracks with ARROW} on two dynamic Objaverse objects, predicted from inputs at 0 and 0.1\,s up to 0.3\,s. The tracks of \method come closest to the ground truth.}
  \label{fig:arrow_tracks}
\end{figure}

\paragraph{Extension to URoPE.}
\label{sec:transfer}
URoPE has no per-token depth, so we lift each match at every depth anchor and move each anchor by its velocity.
\method improves all URoPE metrics on every dataset, generally less than on RayRoPE (\cref{tab:main}).

\paragraph{3D tracking.}
Attached as an adapter to ARROW~\cite{fradlin2026arrow}, a pretrained model for 3D reconstruction and point tracking, \method tracks points more accurately than RayRoPE with a temporal RoPE at every time (\cref{tab:tracking,fig:arrow_tracks}), lowering the 3D end-point error by 10\% between the two input times, 7\% after them, and 4\% at the input times.
A plain adapter of the same size improves on ARROW less, so the gains do not come from the added parameters alone.

\paragraph{Velocities from optical flow.}
On Stereo4D, replacing our matcher with RAFT~\cite{teed2020raft} flow and dropping the matching loss keeps most of the gain: its \psnrd is 0.18\,dB below our matcher's and far above the model without motion (\cref{tab:full}), with slightly better tOF and $\delta_\mathrm{avg}$ but worse JEDi.

\begin{figure}[t]
  \centering
  \includegraphics[width=\linewidth]{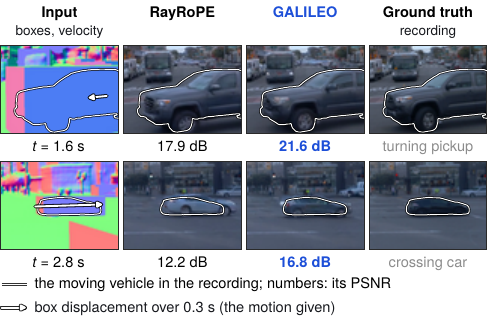}
  \caption{\textbf{A video renderer for code worlds} on held-out Waymo clips, given each vehicle as its 3D box and, for \method, its motion.
  RayRoPE draws the moving vehicle off its place in the recording; \method draws it in place.}
  \label{fig:driving}
\end{figure}

\paragraph{A video renderer for code worlds.}
\method also attaches to a large video generator.
AgentGarten~\cite{mirros2026agentgarten} renders code worlds into video, conditioned on their geometry; we attach its conditioning to Wan2.2-TI2V-5B~\cite{wan2025wan} and generate 4\,s of Waymo~\cite{sun2020waymo} front-camera video from the first frame, per-frame geometry, and vehicle boxes (\cref{sec:supp_agentgarten}).
Moving each key by the velocity of its vehicle's ground-truth box raises \psnrd on moving vehicles over RayRoPE in every second, whereas AgentGarten's token conditioning helps only in the first second (\cref{tab:agentgarten,fig:driving}).
Switching the motion off at test time removes the gain.

\begin{table}[t]
  \centering
  \footnotesize
  \setlength{\tabcolsep}{2.2pt}
  \resizebox{\columnwidth}{!}{%
  \begin{tabular}{@{}l c cccc c@{}}
    \toprule
    & & \multicolumn{4}{c}{second after the first frame} & \\
    \cmidrule(lr){3-6}
    Conditioning & \psnrd & 0--1 & 1--2 & 2--3 & 3--4 & IoU \\
    \midrule
    Token conditioning & 16.98 & 19.73 & 16.90 & 16.07 & 15.70 & .628 \\
    \hspace{0.5em}+ RayRoPE & 17.01 & 19.61 & 17.03 & 16.20 & 15.71 & .660 \\
    \midrule
    \hspace{1em}+ \method & \best{17.45} & \best{20.11} & \best{17.46} & \best{16.76} & \best{16.04} & \best{.673} \\
    \hspace{1em}+ \method, motion off & 16.89 & 19.43 & 16.84 & 16.10 & 15.45 & .630 \\
    \bottomrule
  \end{tabular}}
  \caption{\textbf{A video renderer for code worlds} on Waymo (\cref{sec:supp_agentgarten}): \psnrd on moving vehicles over the 4\,s generated video and per second, and their silhouette IoU (recording: .879). Motion off: the same \method model with every velocity set to zero at test time. Best in bold.}
  \label{tab:agentgarten}
\end{table}

\section{Conclusion}
\label{sec:conclusion}

We introduced \method, which encodes each key where its content is at the query time by transporting it inside the encoding along its motion, and leaves queries and the attention operator unchanged.
Attached to RayRoPE and URoPE, it improves moving content on synthetic and real multi-view videos over encodings that already carry time: attention moves from the static place of the content to the content itself, and the model becomes far less sensitive to the inertial frame of its cameras.
The transport also improves a pretrained 3D tracker and a video generator.

\paragraph{Limitations.}
By default, \method estimates a constant velocity between adjacent frames, which does not follow content that accelerates or turns; 3D tracks or ground-truth motion can replace the estimate where available. %
This estimate has a limited range, and depth and motion accuracy limit the transport (\cref{sec:supp_velocity,sec:where}).
The full model also relies on depth and motion supervision that unlabeled video does not provide.

{
    \small
    \bibliographystyle{ieeenat_fullname}
    \bibliography{sec_new4/main}
}

\clearpage
\iftoggle{cvprfinal}{}{\setcounter{page}{1}}
\renewcommand{\topfraction}{0.9}\renewcommand{\dbltopfraction}{0.9}\renewcommand{\bottomfraction}{0.6}\renewcommand{\textfraction}{0.07}
\renewcommand{\floatpagefraction}{0.8}\renewcommand{\dblfloatpagefraction}{0.8}\setcounter{topnumber}{3}\setcounter{dbltopnumber}{3}\setcounter{bottomnumber}{2}\setcounter{totalnumber}{5}
\maketitlesupplementary
\appendix
\crefalias{section}{appendix}

\section{Method Details}
\label{sec:supp_method}

\paragraph{RayRoPE coordinates and per-view encoding.}
Our coordinate parameterization follows the official RayRoPE implementation~\cite{wu2026rayrope}.
In each transformer layer (indexed from $1$ to $L$), a linear head predicts a log-depth $\ell_b$ and a half-width $\sigma_b$ for each token $b$, defining the depth interval $d_b^{\pm} = \exp(\min(\ell_b \pm \sigma_b, 3))$.
Three rays pass through the patch corners $(0,0)$, $(1,0)$, and $(0,1)$, with interval endpoints along ray $r$ given by
\[ x_{b,r}^{\pm} = c_b + R_b\, d_b^{\pm} \tilde{K}_b^{-1} [u_r, v_r, 1]\T, \]
where $\tilde{K}_b$ denotes the patch-space intrinsics.
For a query token $a$, any 3D point $x$ is projected into the query camera frame via
\[ y = \tilde{K}_a R_a\T (x - c_a). \]
The encoding takes the first two components of the unit direction $y / \lVert y \rVert$ and the clamped depth $\mathrm{clamp}(y_z, -10, 10)$ along each of the three rays, together with the relative camera center $R_a\T (c_b - c_a)$. This gives $3 \cdot 3 + 3 = 12$ coordinates in total.
Because $c_a$, $R_a$, $\tilde{K}_a$, and $t_a$ depend only on the query view rather than on the token's 2D location within that view, the key rotation matrices $\hat{M}_a(b)$ are shared across all queries in a view and computed once per query view.

Each coordinate $\phi_k$ is encoded across six frequencies
\[ \omega_{k,f} = 2\pi \cdot 3^f / T_k, \quad f = 0, \dots, 5, \]
with periods $T_k = 4$ for camera centers, $8$ for ray directions, and $80$ for depth.
This produces $72$ two-dimensional rotation blocks for a head dimension of $144$; when temporal RoPE is enabled, three of these blocks encode time instead (\cref{sec:supp_details}).
Camera-center blocks apply standard rotations $(\cos\theta, \sin\theta)$, while ray blocks integrate rotations analytically over the predicted depth interval, using $C = (\sin\theta^{+} - \sin\theta^{-}) / (\theta^{+} - \theta^{-})$ and $S = (\cos\theta^{-} - \cos\theta^{+}) / (\theta^{+} - \theta^{-})$.
For a query at time $t_a$, \method shifts all six interval endpoints of key $b$ by $v_b\,(t_a - t_b)$ according to \cref{eq:transport} and recomputes the ray blocks, while keeping the camera-center blocks identical to RayRoPE.
In total, \method adds only 243,616 parameters to RayRoPE: 73,728 in the first-stage token-matching projection and 169,888 in the second-stage refinement CNN (\cref{sec:supp_velocity}).

\paragraph{Training losses.}
The total loss of \cref{eq:loss} is
\begin{equation*}
  \mathcal{L} = \mathcal{L}_\text{img} + \mathcal{L}_\text{match} + \mathcal{L}_\text{depth}.
\end{equation*}
$\mathcal{L}_\text{img}$ is the mean squared error plus a perceptual loss with weight $0.5$.
$\mathcal{L}_\text{match}$ adds a token-level term $\mathcal{L}_\text{match}^\text{tok}$ (weight $0.1$) and a pixel-level term $\mathcal{L}_\text{match}^\text{pix}$; both measure $L_1$ correspondence errors (in token units) over visible input tokens against ground-truth matches, weighting dynamic tokens by $5$ in the pixel-refinement stage.
$\mathcal{L}_\text{depth}$ has weight $0.1$ and is the $L_1$ error between the predicted log-depth at every layer and the log of the patch-median ground-truth depth, over all input and target tokens.
On Syn4D and Objaverse, it also weights dynamic input tokens by $5$ and adds two regularizers, on the mean log-depth error per view and on the consistency of the log-depth error between corresponding tokens across time.
On Stereo4D, supervised by stereo depth, it is the $L_1$ term alone, and \method still improves \psnrd over RayRoPE by 1.16\,dB (\cref{tab:main}); on Syn4D, the extra terms change no PSNR metric of RayRoPE by more than 0.10\,dB (\cref{tab:ablations}).
The RayRoPE baseline is trained with the same $\mathcal{L}_\text{depth}$ as \method, whereas URoPE with \method uses no depth loss.
Gradients from $\mathcal{L}_\text{img}$ propagate through the transported coordinates to the pixel-level matcher and, when built on RayRoPE, to the depth heads at both ends of each correspondence; the first-stage token matcher is detached from the second stage and supervised only by $\mathcal{L}_\text{match}^\text{tok}$.

\section{Velocity Estimation}
\label{sec:supp_velocity}

This section details the internal velocity estimator described in \cref{sec:velocity} and ablates its design components on Syn4D (\cref{tab:velocity}, with remaining image metrics in \cref{tab:ablations}).
Starting from the RayRoPE baseline without transport~(a), we progressively add token-level matching across all input frames of the same camera~(b), CNN-based sub-token pixel refinement~(c), the ground-truth velocity curriculum~(d), and restriction of correspondences to temporally adjacent frames~(e), which yields the full \method model.

\begin{table}[!htb]
  \centering
  \footnotesize
  \setlength{\tabcolsep}{2.5pt}
  \resizebox{\columnwidth}{!}{%
  \begin{tabular}{@{}l ccc cc ccc@{}}
    \toprule
    & & & & \multicolumn{2}{c}{image} & \multicolumn{3}{c}{video} \\
    \cmidrule(lr){5-6} \cmidrule(l){7-9}
    & Partners & Refine & Curric. & PSNR$_\mathrm{dyn}\!\uparrow$ & PSNR$\uparrow$ & tOF$\downarrow$ & $\delta_\mathrm{avg}\!\uparrow$ & JEDi$\downarrow$ \\
    \midrule
    (a) & \multicolumn{3}{c}{no motion} & 19.88 & 23.17 & 2.19 & .639 & 1.075 \\
    (b) & all & -- & -- & 20.16\,\up{+0.28} & 23.27\,\up{+0.10} & 2.01\,\up{$-$0.18} & .647\,\up{+.008} & 0.764\,\up{$-$0.310} \\
    (c) & all & \checkmark & -- & 20.14\,\up{+0.25} & 23.37\,\up{+0.20} & 1.90\,\up{$-$0.29} & .656\,\up{+.016} & 0.647\,\up{$-$0.428} \\
    (d) & all & \checkmark & \checkmark & 20.25\,\up{+0.37} & 23.33\,\up{+0.16} & 1.89\,\up{$-$0.30} & .659\,\up{+.019} & 0.593\,\up{$-$0.482} \\
    (e) & adjacent & \checkmark & \checkmark & 20.82\,\up{+0.94} & 23.51\,\up{+0.34} & 1.65\,\up{$-$0.53} & .668\,\up{+.029} & 0.380\,\up{$-$0.695} \\
    \bottomrule
  \end{tabular}}
  \caption{\textbf{Velocity estimation ablations} on Syn4D, using the metrics of \cref{tab:main}.
  Partners: same-camera input frames used for matching (all other frames vs.\ adjacent frames only).
  Refine: CNN-based pixel-level refinement following token matching.
  Curric.: velocity curriculum during early training (\cref{sec:training}).
  Row~(a) corresponds to the RayRoPE baseline in \cref{tab:main} and row~(e) to \method; small numbers denote changes relative to~(a), all statistically significant.
  Remaining image metrics are reported in \cref{tab:ablations}.}
  \label{tab:velocity}
\end{table}
\begin{table*}[t]
  \centering
  \footnotesize
  \setlength{\tabcolsep}{2.5pt}
  \begin{tabular}{l r cc c cc}
    \toprule
    & & \multicolumn{2}{c}{Median error (px)} && \multicolumn{2}{c}{Error} \\
    \cmidrule(lr){3-4} \cmidrule(lr){6-7}
    Tokens & Count & Match & Zero motion & Match better & $\le 4$\,px & $\le 8$\,px \\
    \midrule
    Dynamic, all & 143,580 & 1.18 & 4.21 & 81.1\% & 77.7\% & 86.5\% \\
    \quad visible & 130,159 & 1.05 & 3.99 & 83.5\% & 82.0\% & 90.1\% \\
    \quad\quad in range, textured & 98,926 & 0.95 & 3.71 & 84.5\% & 85.0\% & 92.2\% \\
    Static, visible & 934,464 & 0.36 & 1.94 & 84.1\% & 95.2\% & 97.2\% \\
    \midrule
    \textit{Dynamic tokens by case} & \textit{share} \\
    \quad out of frame & 2.9\% & 20.06 & 21.44 &&& 23.8\% \\
    \quad occluded & 6.4\% & 4.46 & 4.91 &&& 64.9\% \\
    \quad beyond search range & 1.1\% & 53.03 & 52.82 &&& 0.1\% \\
    \quad textureless & 20.6\% & 1.41 & 4.68 &&& 87.9\% \\
    \bottomrule
  \end{tabular}
  \caption{\textbf{Correspondence accuracy} of \method on Syn4D across 100 test inputs (the first 50 of the novel-time and novel-view interpolation sets), totaling 1,228,800 pairs of an input token and an adjacent partner frame.
  Error is measured in pixels at $256 \times 256$ resolution ($1\text{ token} = 8$\,px) between the refined pixel-stage match and the ground-truth correspondence; \emph{Zero motion} leaves each token at its original 2D coordinates.
  Match better: fraction of tokens for which the predicted match is closer to the ground truth than zero motion.
  A token is classified as dynamic if over half of its pixels move by more than $3 \times 10^{-3}$ normalized 3D units, and static if none do.
  Visible: ground-truth correspondence lies within the partner frame and agrees with its ground-truth depth; in range: displacement within $\pm 4.5$ tokens along $x$ and $y$; textured: patch grayscale standard deviation of at least $0.01$.
  The lower block assigns each dynamic token to the first applicable failure category; the remaining 68.9\% are visible, in range, and textured.}
  \label{tab:accuracy}
\end{table*}

\paragraph{Partner-frame selection.}
Each input token is matched only to the temporally adjacent input frames of the same camera (one or two partner frames). Ego-camera motion is accounted for later, during 3D lifting, using the known camera poses.
Temporally adjacent frames from the same camera look similar and are only a short time apart, which makes visual correspondence more reliable and the constant-velocity assumption more accurate.
Restricting matching to adjacent frames~(e) rather than all frames of the same camera~(d) improves \psnrd by 0.57\,dB, and every other metric in \cref{tab:velocity} improves as well.
Matching across wider intervals (200--300\,ms) is more prone to occlusion and appearance changes.

\paragraph{Coarse-to-fine matching.}
Because displacements between adjacent timestamps are often smaller than a single patch, we use a two-stage coarse-to-fine matcher with one differentiable soft-argmax update per stage:
\begin{equation}
  m \leftarrow m + \sum_{\delta} \softmax_{\delta}\!\big(\langle f_{n_b}(p_b),\, f_n(m + \delta) \rangle / \tau\big)\, \delta,
  \label{eq:softargmax}
\end{equation}
where $f_n$ denotes the feature map of frame $n$.
\cref{eq:softargmax} compares the feature of token $b$ at its patch center $p_b$ in source frame $n_b$ against candidate features in partner frame $n$ at offsets $\delta$ around the current estimate $m$.
Correspondences are computed once from the features entering layer~2 and shared across layers~2--6, where key transport is applied to input tokens; layer~1 uses the untransported base encoding.

In the first stage, $f_n$ consists of 64-dimensional $L_2$-normalized token projections $\mathrm{normalize}(W_f h_b)$ on the token grid of frame $n$. This stage starts from $m = p_b$, evaluates a $7 \times 7$ token window, and uses temperature $\tau = 0.05$.
In the second stage, pixel-level features are extracted by a lightweight CNN consisting of six $3 \times 3$ convolutional layers (channel widths $32, 64, 64, 64, 64, 64$, each followed by GroupNorm and GELU, with stride~2 in the second layer and dilations~2 and~4 in the fourth and fifth layers) and a final $1 \times 1$ projection.
The CNN takes images scaled to $[-1, 1]$ and produces 32-dimensional $L_2$-normalized feature maps at $128 \times 128$ resolution.
The refinement stage queries the feature at the center pixel of token $b$, initializes $m$ at the detached first-stage match, and searches a $\pm 12$\,px window with a stride of 2\,px and temperature $\tau = 0.02$.
Token-level matching alone~(b) improves \psnrd by 0.28\,dB over~(a); adding pixel refinement~(c) further doubles the gain in full-image PSNR (+0.20 vs.\ +0.10\,dB) and improves all video metrics while leaving \psnrd nearly unchanged.

\paragraph{3D lifting and velocity fitting.}
In each layer, a token's 3D position at its capture time is obtained by lifting its center ray to that layer's predicted depth $e^{\ell_b}$. Its position at each partner time is obtained by lifting the matched pixel with the depth bilinearly sampled from the partner frame's depth map at the same layer (for URoPE, lifting is done independently at each fixed depth anchor).
We then fit a constant 3D velocity $v_b$ by ordinary least squares across the token's own lifted point and its valid partner points $\{x_n\}$ at timestamps $\{t_n\}$:
\begin{equation}
  v_b = \frac{\sum_n (t_n - \bar{t})(x_n - \bar{x})}{\sum_n (t_n - \bar{t})^2},
  \label{eq:fit}
\end{equation}
which reduces to a two-point finite difference for boundary frames (one partner) and a linear regression across three timestamps for interior frames (two partners).
Correspondences falling outside the image boundaries are discarded, and tokens without valid partners are assigned $v_b = 0$.

\paragraph{Velocity curriculum.}
As described in \cref{sec:training}, keys are transported with ground-truth velocities for the first $s_0$ steps, and the velocities are then linearly annealed to the predicted ones by step $s_1$. Each token's ground-truth velocity is computed with \cref{eq:fit} along its ground-truth 3D trajectory over the same timestamps.
We set $(s_0, s_1) = (5\text{k}, 15\text{k})$ for 20k-step runs and $(20\text{k}, 60\text{k})$ for 80k-step runs.
Adding this curriculum~(d) increases \psnrd by 0.11\,dB and further reduces JEDi, with a minor 0.04\,dB trade-off in full-image PSNR.

\paragraph{Correspondence accuracy.}
\cref{tab:accuracy} evaluates the 2D correspondence accuracy of \method on Syn4D against ground-truth adjacent-frame matches at $256 \times 256$ resolution, alongside a zero-motion baseline.
Ground-truth correspondences are obtained by lifting each token's center pixel at its patch-median ground-truth depth, translating it by the median 3D displacement of dynamic pixels within the patch, and projecting it into the partner frame.
Across all dynamic tokens, \method achieves a median error of 1.18\,px compared with 4.21\,px under zero motion, outperforming zero motion on 81.1\% of tokens; on visible static tokens, the median error is 0.36\,px.
Roughly 10\% of dynamic tokens move out of frame, become occluded, or exceed the $\pm 4.5$-token total search range ($\pm 3$ tokens in Stage~1 plus $\pm 12$\,px in Stage~2). For these tokens, the errors are comparable to zero motion.
Another 20.6\% lie in low-texture patches, where the median error rises modestly to 1.41\,px, still well below the 4.68\,px error without matching.
Overall, 2D correspondences are sub-pixel to near-pixel accurate for most dynamic tokens. The residual 3D velocity errors still leave room for a further +0.44\,dB in \psnrd, which is the gain obtained with ground-truth motion and predicted depth (\cref{tab:motion}).

\section{Experimental Details}
\label{sec:supp_details}

\paragraph{Syn4D.}
We use Syn4D~V1 (CC~BY~4.0)~\cite{jiang2026syn4d}, which contains 29 synthetic scenes (289 sequences) with a BEDLAM2~\cite{tesch2025bedlam2} human subject and three animated Objaverse objects, captured by eight synchronized moving cameras at 30\,fps.
All frames are resized to $256 \times 256$.
Every seventh scene in sorted order (\emph{arena}, \emph{desert bald}, \emph{lab bald}, \emph{soviet group}, and \emph{winter group static}) is held out for testing, while the first 128 frames of sequences in the remaining 24 scenes are used for training.
Each input consists of two cameras with optical axes within $30^\circ$ of one another, observed across four timestamps spaced three frames (100\,ms) apart.
During training, each sample supervises a single target frame drawn either within the input window or one frame after the final input.
With probability $0.5$, we select a rig camera positioned between the two input views, defaulting to an input camera when no intermediate rig camera exists; novel-view targets thus constitute roughly 30\% of training samples.
Evaluation splits and counts are summarized in \cref{tab:sets}.

\begin{table}[!htb]
  \centering
  \footnotesize
  \setlength{\tabcolsep}{2pt}
  \begin{tabular}{lllccc}
    \toprule
    Split & Camera & Target time & Seq. & Inputs & Targets \\
    \midrule
    Novel time, interp & input & within window & 100 & 100 & 200 \\
    Novel time, extrap & input & +33\,ms after inputs & 100 & 100 & 200 \\
    Novel view, obs & near & at input times & 43 & 286 & 286 \\
    Novel view, interp & near & within window & 43 & 286 & 286 \\
    Novel view, extrap & near & +33\,ms after inputs & 43 & 286 & 286 \\
    \bottomrule
  \end{tabular}
  \caption{\textbf{Syn4D evaluation splits.}
  Each input provides two cameras at frames $0, 3, 6,$ and $9$; interpolation targets are sampled from frames $\{1, 2, 4, 5, 7, 8\}$ and extrapolation targets at frame $10$.
  Novel-time inputs evaluate one target per input camera.
  Novel-view targets use rig cameras whose optical axes lie within $5^\circ$ of an input camera and whose centers lie within $0.03$--$0.3$ normalized units of that camera's center (median $3.5^\circ$ and $0.10$, where median input depth is $1$), with one target per input.}
  \label{tab:sets}
\end{table}

\paragraph{Dynamic Objaverse.}
We render animated Objaverse~\cite{deitke2023objaverse} assets in Blender using six cameras per object, training on 3,607 objects and evaluating on 197 disjoint test objects across all protocols.
Each input consists of two cameras $15\text{--}30^\circ$ apart, observed at four timestamps spaced 0.1\,s apart.
Novel-time targets query the input cameras at an interior timestamp within each interval (interpolation) and $1/30$\,s after the final input (extrapolation).
Novel-view targets query the remaining four cameras at the input timestamps (\emph{obs}) as well as at the interpolated and extrapolated timestamps. Two of these cameras lie at one-third and two-thirds of the arc between the input views. The other two are outside viewpoints that extend the arc by $10\text{--}25^\circ$ beyond each endpoint, rotated by up to $45^\circ$ off the arc, with independently sampled focal lengths and distances.
Outside viewpoints are never seen during training, and in-between and outside novel views are weighted equally at evaluation.

\paragraph{Stereo4D.}
We use synchronized left and right video streams from Stereo4D~\cite{jin2025stereo4d} ($512 \times 512$, $60^\circ$ field of view, 30\,fps) together with their calibrated camera poses, resizing 30-frame clips to $256 \times 256$.
Metric stereo depth is obtained from left--right disparity estimated by RAFT~\cite{teed2020raft} at $512 \times 512$ resolution.
Inputs consist of both stereo views at four timestamps spaced five frames apart; interpolation targets lie between input frames, and extrapolation targets lie two frames after the last input.
We train on 15,000 clips and evaluate on 600 clips from held-out videos, with one interpolation and one extrapolation target per clip.
Because Stereo4D lacks ground-truth 3D scene flow, dynamic pixels are identified where RAFT optical flow deviates from static depth reprojection by more than 1.5\,px (matching the threshold used on Syn4D), covering roughly 2\% of image pixels.
Similarly, 3D tracks estimated by ARROW~\cite{fradlin2026arrow} provide pseudo-ground-truth correspondences and velocities for $\mathcal{L}_\text{match}$ and the training curriculum.
Stereo4D models omit the dynamic weighting and the two regularizers of the depth loss (\cref{sec:supp_method}).

\paragraph{Dynamic masks and image metrics.}
Each scene is normalized by placing the first input camera at the origin and scaling coordinates such that the median input depth equals $1$.
On Syn4D and Objaverse, we project each foreground target pixel's ground-truth 3D surface point at the target time and at the nearest input time into the target camera; a pixel is marked dynamic if the two projections differ by more than 1.5\,px at $256 \times 256$ resolution.
On Syn4D, the foreground is all pixels with valid depth (excluding sky and background beyond 300\,m), and the motion comes from the human subject and the animated objects. Dynamic pixels are absent at input times (\emph{obs}) and cover 4.9--6.1\% of pixels otherwise.
For each test sample, \psnrd is computed from the mean squared error pooled across dynamic pixels of all its targets, while \ssimd and \lpipsd average the spatial SSIM and LPIPS maps over the same mask (omitting samples with empty dynamic masks).
Full-image PSNR is likewise pooled across each sample's targets and averaged over samples.
As a check, we also use an alternative 3D criterion on Syn4D: a pixel is marked dynamic if its 3D surface point moves by more than $3 \times 10^{-3}$ normalized units ($0.3\%$ of median depth) relative to an input timestamp. Under this criterion, \method still improves over RayRoPE by 0.73\,dB in \psnrd.
Across all tables, image metrics are first averaged over the protocols within each task, and novel views and novel times then receive equal weight. On Stereo4D, interpolation and extrapolation are weighted equally; on Objaverse, in-between and outside novel views are averaged equally first. Novel-view full-image metrics include \emph{obs}.

\paragraph{Video metrics and statistical testing.}
Video evaluations cover 286 novel-view and 200 novel-time clips on Syn4D, 394 novel-time and 788 novel-view 11-frame clips on Objaverse (394 in-between and 394 outside across the 197 test objects), and 1,200 clips on Stereo4D (600 segments rendered for both eyes).
Frames coinciding with input timestamps are excluded from per-frame video metrics, and novel views and novel times receive equal weight.
tOF~\cite{chu2020tecogan} measures the $L_1$ endpoint difference between RAFT~\cite{teed2020raft} optical flows of consecutive rendered frames and consecutive ground-truth frames over dynamic regions.
For $\delta_\mathrm{avg}$~\cite{doersch2022tapvid}, we track ground-truth moving points on the rendered videos with CoTracker3~\cite{karaev2024cotracker3} and average the fraction of tracks within $\{1, 2, 4, 8, 16\}$\,px of the reference trajectories.
Reference trajectories are obtained by projecting ground-truth 3D surface trajectories into the target camera (visibility from ground-truth depth on Syn4D and from Blender surface renders on Objaverse); on Stereo4D, reference tracks are extracted by CoTracker3 on ground-truth videos for query points identified via static stereo reprojection.
JEDi~\cite{luo2025jedi} computes distributional distances over V-JEPA ViT-H/16 features~\cite{bardes2024vjepa} using the Something-Something-v2 attentive probe pooler~\cite{luo2025jedi}. Compared with I3D-based FVD~\cite{unterthiner2018fvd}, it is more sensitive to temporal artifacts and more sample-efficient.
Confidence intervals are 90\% bootstrap intervals resampled over sequences, clips, or objects, and JEDi differences are tested with paired permutation tests at $p < 0.05$.

\paragraph{Backbone and training.}
All view-synthesis models share a decoder-only LVSM~\cite{jin2025lvsm} architecture with six pre-norm transformer layers of hidden dimension $1152$, eight attention heads, FFN dimension $1024$, and query--key RMSNorm~\cite{zhang2019rmsnorm,henry2020qknorm}.
Images are tokenized into $8 \times 8$ patches; input tokens concatenate RGB values with camera-frame Pl\"ucker rays, while target tokens take camera-frame Pl\"ucker rays alone (the Pl\"ucker baseline uses world-frame rays instead; see below).
Optimization follows the official RayRoPE training setup~\cite{wu2026rayrope} (AdamW, learning rate $10^{-4}$, weight decay 0.05, fp16 mixed precision) with batch size $12$ for 20k steps on a single GPU unless noted otherwise.
Smaller LVSM experiments were run on NVIDIA RTX 4090 GPUs and all other experiments, including the ARROW tracking adaptation and the AgentGarten renderer, on NVIDIA B300 GPUs.
Ground-truth depths and correspondences are used only for the training losses and the early-training curriculum; no privileged geometry or motion is used at inference.
\method adds 243,616 parameters (+0.5\%) to the 47.37M-parameter RayRoPE model; \cref{tab:cost} reports the computational overhead for the 12-layer models.
Across three random seeds on Syn4D, the RayRoPE baseline has a standard deviation of at most 0.06\,dB in PSNR, \psnrd, and static-region PSNR.

\paragraph{Baselines.}
The Pl\"ucker baseline uses no rotary encoding: tokens carry per-pixel Pl\"ucker coordinates $(d, o \times d)$ in the normalized world frame together with a zero-initialized linear projection of sinusoidal Fourier features of $t - t_\text{target}$ at periods $16 / 2^k$\,s ($k = 0, \dots, 7$).
On Objaverse, this baseline failed to converge across all five trials, with validation metrics stagnating between 5k and 20k steps.
Following standard practice in video transformers~\cite{wang2024qwen2vl,yang2025cogvideox}, temporal RoPE allocates dedicated rotary planes to time, rotating keys and queries at angular frequency $\omega = 2\pi$\,rad/s (period 1\,s) so that attention depends on the relative offset $t_a - t_b$.
In RayRoPE, temporal RoPE replaces the three highest-frequency ($f = 5$) camera-center blocks, leaving 69 of the 72 blocks for spatial geometry; RayRoPE and RayRoPE~+~\method use the same configuration on all datasets.
PRoPE~\cite{li2025prope} and GTA~\cite{miyato2024gta} operate on camera-ray tokens, and URoPE~\cite{xie2026urope} follows its official configuration with depth anchors at $\{0.4, 1.2, 2.0, 2.8\}$.

\paragraph{URoPE with \method.}
Because URoPE does not predict per-token depths, we lift each 2D correspondence at all four depth anchors, fit an independent 3D velocity per anchor with \cref{eq:fit}, and transport each anchor point by its own velocity.
The anchor points generally lie off the surface, so ground-truth 3D surface velocities do not correspond to them; the training curriculum therefore substitutes ground-truth 2D correspondences lifted at each anchor depth.
Unlike RayRoPE~+~\method, URoPE~+~\method detaches the backbone token features before first-stage matching, so that $\mathcal{L}_\text{match}^\text{tok}$ updates only the matching projection.
As shown in \cref{tab:detach}, detaching and not detaching perform comparably across all metrics (+0.49 vs.\ +0.44\,dB in \psnrd).

\section{Attention Analysis}
\label{sec:supp_attention}

This section provides full experimental details for the attention analysis presented in \cref{sec:attention}.

\paragraph{Clip and frame selection.}
From the 600 Stereo4D test clips, we identify the 220 clips with over 200 dynamic pixels in at least six held-out left-camera frames, randomly draw 80 clips, and select one held-out left-camera frame per clip, namely the one with the largest dynamic region among the frames two steps from the nearest input timestamp.
Inputs consist of both stereo views at all four input timestamps, as in standard evaluation.

\paragraph{Attention computation.}
At each transformer layer, we extract the rotary-encoded target queries and the rotary-encoded keys of all tokens to compute the attention weights $\softmax(q\T k / \sqrt{d})$.
For a given target token and each of the eight input views, we define the \emph{share of a frame's attention} as the sum of attention weights assigned to a subset of tokens within that view divided by the total attention received by that view.
Scores are averaged across all eight attention heads and, unless noted otherwise, across layers~2--6 (the five layers where key transport is active).

\paragraph{Content and static locations.}
For each target token and input view, the \emph{content place} is determined with bidirectional RAFT~\cite{teed2020raft} optical flow computed at $2\times$ resolution between the ground-truth target frame and the input frame.
A token's correspondence is its patch center shifted by the median flow of its 64 pixels; it is marked valid if the median forward--backward consistency error is below 1.5\,px and the matched point lies within the image bounds.
The \emph{static place} is obtained by lifting the target token's center at its patch-median ground-truth stereo depth and projecting this 3D point into the input view under a static-scene assumption; it is marked valid when more than half of the patch pixels have valid stereo depth.
Both reference locations are computed independently of one another and of the evaluated models.
On purely static tokens, the two locations agree closely up to flow and stereo estimation noise, with a median discrepancy of 0.24\,px (90th percentile: 1.13\,px) across 353,619 token--view pairs.

\paragraph{Pair selection and aggregation.}
A target-token and input-view pair is included in the moving-content evaluation if at least half of the token's pixels are dynamic, both reference locations are valid, and the two locations are separated by at least 10\,px so that they are resolvable on the $8 \times 8$ patch grid.
Excluding clips with fewer than eight qualifying pairs yields 4,587 pairs across 46 clips.
Attention at each location is measured as the share of the view's attention falling on tokens whose centers lie within a disc whose radius is half the separation between the two locations, so that the two regions do not overlap and have equal area.
For static content, we evaluate tokens containing no dynamic pixels with valid reference locations, measuring the share of attention within 12\,px of the content place.
The 1D profile in \cref{fig:attention} (middle) projects token centers onto the axis connecting the static and content places (normalized by their separation) and aggregates attention within a strip of half-width $0.5$ in bins of width $0.2$.
Reported statistics average across pairs, with 95\% bootstrap confidence intervals computed over clips.

\begin{table}[!htb]
  \centering
  \footnotesize
  \setlength{\tabcolsep}{4pt}
  \begin{tabular}{l cc}
    \toprule
    Share of a frame's attention (\%) & RayRoPE & \method \\
    \midrule
    \emph{Moving content} \\
    \quad at the content place & 6.9 {\scriptsize [4.9, 9.7]} & \best{22.2} {\scriptsize [18.4, 24.9]} \\
    \quad at the static place & 20.8 {\scriptsize [17.5, 23.8]} & 8.8 {\scriptsize [6.7, 11.8]} \\
    \emph{Static content} \\
    \quad within 12\,px of content & 45.7 & 45.0 \\
    \bottomrule
  \end{tabular}
  \caption{\textbf{Attention allocation} on Stereo4D (4,587 moving token--view pairs across 46 clips; averaged over heads and layers~2--6, with 95\% clip-bootstrap confidence intervals in brackets).}
  \label{tab:attention}
\end{table}

\begin{figure}[!htb]
  \centering
  \includegraphics[width=\linewidth]{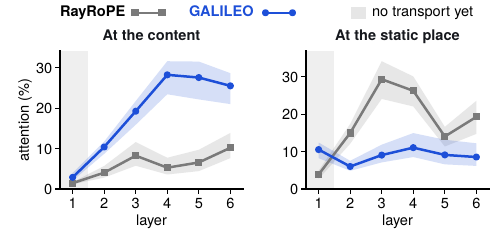}
  \caption{\textbf{Layer-wise attention on moving patches} (using the pairs and definitions of \cref{fig:attention}). Layer~1 of \method runs before velocity estimation and has no key transport (shaded gray); both models place little attention on moving content there. From layer~2 onward, \method shifts attention strongly toward the content place, whereas RayRoPE concentrates attention on the static place.}
  \label{fig:layers}
\end{figure}
\begin{figure}[!htb]
  \centering
  \includegraphics[width=\linewidth]{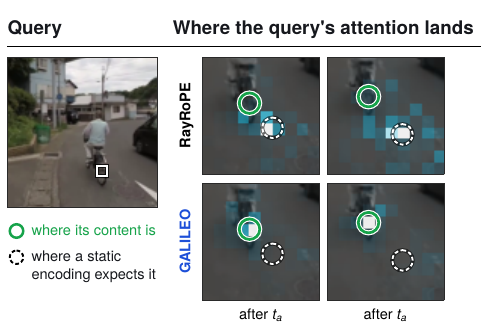}
  \caption{\textbf{Attention follows the encoding} on an additional Stereo4D clip: a query patch on the moving cyclist at held-out time $t_a$ and its attention maps in the two temporally farthest input frames, following the conventions of \cref{fig:attention}; in this clip both lie after $t_a$.}
  \label{fig:attention_cyclist}
\end{figure}

\paragraph{Results.}
\cref{tab:attention} reports the aggregate attention shares corresponding to \cref{fig:attention}, and \cref{fig:layers} breaks them down by layer.
Layer~1 of \method applies no transport and thus serves as an internal control within the same network. It allocates 2.9\% of attention to the content place versus 10.5\% to the static place, the same static bias that RayRoPE shows in all layers.
From layer~2 onward, where key transport is active, \method's attention at the content place rises to 26--28\% in layers~4--6 while staying between 6\% and 11\% at the static place. In contrast, RayRoPE allocates 14--29\% to the static place and 4--10\% to the content place.
Across the 46 held-out frames, the percentage-point increase in content-place attention under \method correlates with the per-frame gain in \psnrd (Pearson $r = 0.67$).
\cref{fig:attention_cyclist} shows an additional example on a second Stereo4D clip.

\section{Galilean Boost Evaluation}
\label{sec:supp_boost}

\paragraph{Protocol.}
We evaluate reference-frame robustness on the 80 held-out Stereo4D frames of \cref{sec:supp_attention}.
In normalized scene coordinates (first input camera at the origin at $t = 0$, median input depth $1$), every camera center $c_n$ across the eight input views and the target view is shifted to $c_n + w\,t_n$, while rotations, intrinsics, timestamps, input images, and ground-truth target images remain unchanged.
We set $w = p / (f\,T)$ along the horizontal axis of the first camera, with focal length $f$ (in pixels) and input span $T = 0.5$\,s, producing an apparent drift of $p$ pixels at the median depth over the input window.
Static-region PSNR is evaluated on the complement of the dynamic mask dilated by 8\,px.

\begin{table}[!htb]
  \centering
  \footnotesize
  \setlength{\tabcolsep}{4pt}
  \begin{tabular}{ll ccccc}
    \toprule
    & & \multicolumn{5}{c}{Drift over the input window (px)} \\
    \cmidrule(lr){3-7}
    Region & Method & 0 & 4 & 8 & 16 & 32 \\
    \midrule
    \multirow{2}{*}{Static} & RayRoPE & 30.95 & 28.95 & 26.71 & 23.72 & 20.92 \\
    & \method & \best{31.07} & \best{30.83} & \best{30.55} & \best{29.74} & \best{27.60} \\
    \midrule
    \multirow{2}{*}{Moving} & RayRoPE & 24.53 & 23.34 & 22.48 & 21.16 & 19.46 \\
    & \method & \best{26.07} & \best{25.84} & \best{25.73} & \best{25.34} & \best{24.11} \\
    \midrule
    \multirow{2}{*}{Full image} & RayRoPE & 28.63 & 27.11 & 25.45 & 22.91 & 20.39 \\
    & \method & \best{29.34} & \best{29.14} & \best{28.93} & \best{28.33} & \best{26.45} \\
    \bottomrule
  \end{tabular}
  \caption{\textbf{Robustness to Galilean boosts}: PSNR (dB) as a function of reference-frame speed across 80 held-out Stereo4D frames. A drift of 0\,px corresponds to the original capture frame.}
  \label{tab:boost}
\end{table}
\begin{figure*}[t]
  \centering
  \includegraphics[width=\textwidth]{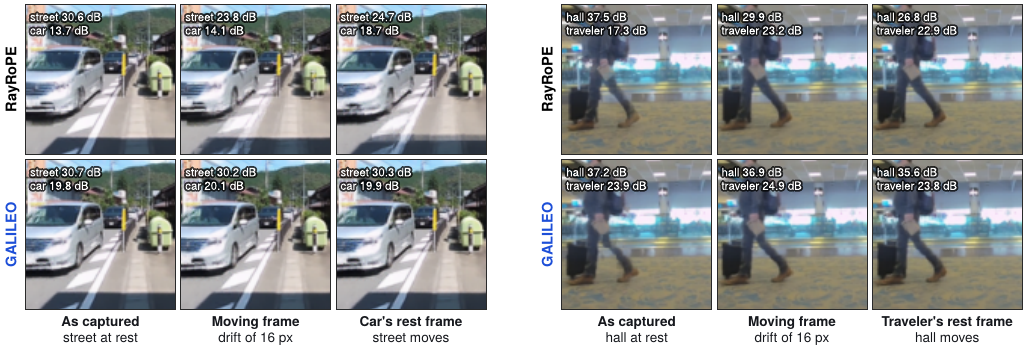}
  \caption{\textbf{The Galilean boost test} on two more Stereo4D clips (renders as in \cref{fig:boost}, top; numbers: PSNR of the static surroundings and of the moving object). In the moving object's rest frame, RayRoPE renders the object better and its static surroundings much worse, whereas \method renders both well in every frame.}
  \label{fig:boost_supp}
\end{figure*}

\paragraph{Results.}
\cref{tab:boost} reports PSNR across static regions, moving regions, and full images as a function of reference-frame drift.
In a moving frame, all scene content has a nonzero coordinate velocity, and RayRoPE degrades sharply across the entire image: at a 16\,px drift, PSNR drops by 7.2\,dB on static regions and 3.4\,dB on moving regions, compared with only 1.3\,dB and 0.7\,dB for \method.
The same pattern holds for a vertical boost: repeating the sweep with $w$ along the camera's vertical axis on 23 test clips lowers static-region PSNR for RayRoPE by 5.8, 9.2, and 11.8\,dB at drifts of 8, 16, and 32\,px, versus 0.2, 0.6, and 3.6\,dB for \method.
Since the underlying visual observations are identical at every boost speed, this sweep also measures, in a controlled way, how the error of a static encoding scales with the apparent object speed.

\paragraph{Object rest-frame transformation.}
In \cref{fig:boost} (top) and \cref{fig:boost_supp}, we set $w$ to minus the foreground velocity, estimated as the median 3D velocity predicted by \method over input tokens corresponding to the target's dynamic region; this brings the moving object to rest while setting the background into uniform motion.
On the train sequence (\cref{fig:boost}), shifting into the train's rest frame improves RayRoPE by 5.2\,dB on the train (20.4 to 25.6\,dB) while degrading the platform by 3.8\,dB (35.8 to 32.0\,dB), whereas \method varies by only +0.6 and $-$0.9\,dB, respectively.
On the car and traveler sequences (\cref{fig:boost_supp}), the same role reversal occurs under RayRoPE: in the object's rest frame, the car rises from 13.7 to 18.7\,dB and the traveler from 17.3 to 22.9\,dB, while the street drops from 30.6 to 24.7\,dB and the hall from 37.5 to 26.8\,dB.
In contrast, \method renders the car at 19.8 and 19.9\,dB (original and rest frame, respectively), the traveler at 23.9 and 23.8\,dB, the street at 30.7 and 30.3\,dB, and the hall at 37.2 and 35.6\,dB.

\section{Complete Image Metrics}
\label{sec:supp_protocol}
\begin{table*}[t]
  \centering
  \footnotesize
  \setlength{\tabcolsep}{5pt}
  \begin{tabular}{l ccc c ccc}
    \toprule
    & \multicolumn{3}{c}{dynamic region} & static & \multicolumn{3}{c}{full image} \\
    \cmidrule(lr){2-4} \cmidrule(lr){5-5} \cmidrule(l){6-8}
    Method & PSNR$\uparrow$ & SSIM$\uparrow$ & LPIPS$\downarrow$ & PSNR$\uparrow$ & PSNR$\uparrow$ & SSIM$\uparrow$ & LPIPS$\downarrow$ \\
    \midrule
    \multicolumn{8}{l}{\textit{Syn4D}} \\
    Pl\"ucker & 19.20 & .443 & .373 & 21.93 & 21.72 & .578 & .322 \\
    PRoPE & 19.34 & .467 & .414 & 22.86 & 22.57 & .612 & .364 \\
    GTA & 19.28 & .462 & .413 & 22.76 & 22.46 & .610 & .366 \\
    URoPE & 19.66 & .481 & .315 & 23.32 & 22.99 & .634 & .258 \\
    \quad + \method & 20.14 & .504 & .316 & 23.38 & 23.11 & .637 & .266 \\
    RayRoPE & 19.88 & .487 & .268 & 23.50 & 23.17 & .646 & .226 \\
    \quad + \method & \textbf{20.82} & \textbf{.535} & \textbf{.256} & \textbf{23.75} & \textbf{23.51} & \textbf{.653} & \textbf{.217} \\
    \midrule
    \multicolumn{8}{l}{\textit{Dynamic Objaverse}} \\
    PRoPE & 16.73 & .470 & .2373 & 21.04 & 27.51 & .9505 & .0776 \\
    GTA & 16.81 & .471 & .1779 & 21.59 & 27.76 & .9557 & .0538 \\
    URoPE & 16.33 & .462 & .2380 & 20.38 & 27.07 & .9533 & .0735 \\
    \quad + \method & 16.70 & .474 & .2362 & 20.65 & 27.25 & .9538 & .0738 \\
    RayRoPE & 16.67 & .461 & .1885 & 20.94 & 27.42 & .9545 & .0549 \\
    \quad + \method & \textbf{17.96} & \textbf{.521} & \textbf{.1622} & \textbf{21.64} & \textbf{28.25} & \textbf{.9600} & \textbf{.0466} \\
    \midrule
    \multicolumn{8}{l}{\textit{Stereo4D}} \\
    Pl\"ucker & 23.59 & .619 & .0839 & 29.44 & 29.17 & .8593 & .0388 \\
    PRoPE & 24.06 & .645 & .0819 & 30.22 & 29.95 & .8754 & .0354 \\
    GTA & 24.15 & .647 & .0823 & 30.34 & 30.07 & .8785 & .0355 \\
    URoPE & 24.61 & .666 & .0770 & 31.04 & 30.73 & .8929 & .0323 \\
    \quad + \method & 25.41 & .707 & .0736 & 31.24 & 31.00 & .8987 & .0313 \\
    RayRoPE & 24.61 & .666 & .0754 & 31.38 & 31.04 & .9071 & .0295 \\
    \quad + \method & \textbf{25.78} & \textbf{.728} & \textbf{.0729} & \textbf{31.58} & \textbf{31.34} & \textbf{.9105} & \textbf{.0294} \\
    \quad + \method, RAFT flow & 25.60 & .725 & .0713 & 31.68 & 31.43 & .9137 & .0286 \\
    \bottomrule
  \end{tabular}
  \caption{\textbf{Complete image metrics} at 20k steps, averaged across protocols as in \cref{tab:main} (\cref{sec:supp_details}).
  Static PSNR is computed on the complement of the dynamic region (within the object foreground on Objaverse), and novel-view full-image metrics include input-time targets (\emph{obs}).
  All encodings use temporal RoPE (kept in \method) except Pl\"ucker, which uses a Fourier time embedding; RayRoPE uses the same depth losses as \method (\cref{sec:supp_method}), and URoPE~+~\method detaches first-stage matching features (\cref{tab:detach}).
  Pl\"ucker failed to converge on Objaverse (4.91\,dB \psnrd, 16.86\,dB PSNR) and is omitted.
  The final row replaces our internal matcher with external RAFT~\cite{teed2020raft} optical flow and drops $\mathcal{L}_\text{match}$; it is excluded from bolding.
  Best results among self-contained models are in bold.}
  \label{tab:full}
\end{table*}

\cref{tab:full} reports the complete set of dynamic-region, static-region, and full-image metrics corresponding to \cref{tab:main}.

\paragraph{Syn4D and Dynamic Objaverse.}
On both synthetic benchmarks, RayRoPE~+~\method is best on every dynamic, static, and full-image metric (on Objaverse, it beats GTA by only 0.05\,dB in static PSNR).
Paired with URoPE on Syn4D, \method improves \psnrd and \ssimd while leaving \lpipsd virtually unchanged (.315 vs.\ .316) and slightly worsening full-image LPIPS (.258 to .266).

\paragraph{Stereo4D.}
On real-world stereo videos, \method improves every metric on both URoPE and RayRoPE (with full-image LPIPS on RayRoPE changing marginally from .0295 to .0294).
Replacing our internal matcher with external RAFT~\cite{teed2020raft} optical flow retains most of the dynamic-region gain (25.60 vs.\ 25.78\,dB in \psnrd), and its static-region and full-image PSNR are slightly higher.
On the videos it has slightly better tOF (0.75 vs.\ 0.80) and $\delta_\mathrm{avg}$ (.963 vs.\ .958) but worse JEDi (0.363 vs.\ 0.343); all three differences are beyond noise.

\section{Additional Ablation Studies}
\label{sec:supp_ablations}

\begin{table*}[t]
  \centering
  \footnotesize
  \setlength{\tabcolsep}{4pt}
  \begin{tabular}{@{}l ccc c cc@{}}
    \toprule
    & \multicolumn{3}{c}{dynamic region} & static & \multicolumn{2}{c}{full image} \\
    \cmidrule(lr){2-4} \cmidrule(lr){5-5} \cmidrule(l){6-7}
    Variant & PSNR$\uparrow$ & SSIM$\uparrow$ & LPIPS$\downarrow$ & PSNR$\uparrow$ & PSNR$\uparrow$ & LPIPS$\downarrow$ \\
    \midrule
    \multicolumn{7}{@{}l}{\textit{Velocity estimation} ($\Delta$ vs.\ (a))} \\
    \quad RayRoPE, basic depth loss only & 19.84\,\nn{$-$0.04} & .485\,\nn{$-$.002} & .272\,\dn{+.004} & 23.59\,\up{+0.10} & 23.26\,\up{+0.09} & .224\,\up{$-$.002} \\
    \quad (a) RayRoPE + depth losses & 19.88 & .487 & .268 & 23.50 & 23.17 & .226 \\
    \quad (b) token matching & 20.16\,\up{+0.28} & .498\,\up{+.010} & .264\,\up{$-$.004} & 23.58\,\up{+0.08} & 23.27\,\up{+0.10} & .223\,\up{$-$.002} \\
    \quad (c) + pixel refinement & 20.14\,\up{+0.25} & .498\,\up{+.010} & .262\,\up{$-$.006} & 23.68\,\up{+0.18} & 23.37\,\up{+0.20} & .219\,\up{$-$.007} \\
    \quad (d) + curriculum & 20.25\,\up{+0.37} & .505\,\up{+.018} & .259\,\up{$-$.009} & 23.62\,\up{+0.13} & 23.33\,\up{+0.16} & .219\,\up{$-$.007} \\
    \quad (e) + adjacent only (\method) & 20.82\,\up{+0.94} & .535\,\up{+.047} & .256\,\up{$-$.012} & 23.75\,\up{+0.25} & 23.51\,\up{+0.34} & .217\,\up{$-$.009} \\
    \midrule
    \multicolumn{7}{@{}l}{\textit{Supervision and motion without transport} ($\Delta$ vs.\ (a))} \\
    \quad image loss only & 20.44\,\up{+0.56} & .514\,\up{+.027} & .282\,\dn{+.014} & 23.57\,\up{+0.07} & 23.31\,\up{+0.14} & .235\,\dn{+.009} \\
    \quad velocities as token features & 20.10\,\up{+0.22} & .493\,\up{+.005} & .262\,\up{$-$.006} & 23.70\,\up{+0.20} & 23.38\,\up{+0.21} & .219\,\up{$-$.007} \\
    \quad matching losses, no key transport & 20.11\,\up{+0.23} & .492\,\up{+.004} & .263\,\up{$-$.005} & 23.70\,\up{+0.21} & 23.38\,\up{+0.21} & .218\,\up{$-$.008} \\
    \midrule
    \multicolumn{7}{@{}l}{\textit{Oracles} ($\Delta$ vs.\ (a))} \\
    \quad GT motion$^\dagger$ & 21.26\,\up{+1.38} & .560\,\up{+.073} & .241\,\up{$-$.028} & 23.84\,\up{+0.35} & 23.66\,\up{+0.49} & .210\,\up{$-$.016} \\
    \quad GT depth$^\ddagger$ & 21.43\,\up{+1.55} & .566\,\up{+.078} & .177\,\up{$-$.091} & 25.46\,\up{+1.96} & 25.09\,\up{+1.92} & .153\,\up{$-$.073} \\
    \quad GT motion + GT depth & 23.75\,\up{+3.87} & .692\,\up{+.204} & .145\,\up{$-$.123} & 25.79\,\up{+2.30} & 25.66\,\up{+2.49} & .142\,\up{$-$.084} \\
    \midrule
    \multicolumn{7}{@{}l}{\textit{Warping instead of transport} ($\Delta$ vs.\ GT motion + GT depth; predicted warp vs.\ (e))} \\
    \quad GT warp & 23.85\,\nn{+0.11} & .674\,\dn{$-$.017} & .166\,\dn{+.021} & 25.41\,\dn{$-$0.39} & 25.30\,\dn{$-$0.36} & .151\,\dn{+.009} \\
    \quad predicted warp & 20.58\,\dn{$-$0.24} & .525\,\dn{$-$.009} & .261\,\dn{+.005} & 23.49\,\dn{$-$0.26} & 23.26\,\dn{$-$0.25} & .210\,\up{$-$.007} \\
    \midrule
    \multicolumn{7}{@{}l}{\textit{Motion off at test} ($\Delta$ vs.\ the model with motion)} \\
    \quad \method & 19.84\,\dn{$-$0.98} & .481\,\dn{$-$.054} & .264\,\dn{+.008} & 23.63\,\dn{$-$0.12} & 23.29\,\dn{$-$0.22} & .218\,\nn{+.001} \\
    \quad GT motion + GT depth & 21.06\,\dn{$-$2.69} & .542\,\dn{$-$.150} & .185\,\dn{+.040} & 25.28\,\dn{$-$0.52} & 24.91\,\dn{$-$0.75} & .153\,\dn{+.010} \\
    \bottomrule
  \end{tabular}
  \caption{\textbf{Additional ablations and oracles} on Syn4D at 20k steps (all with temporal RoPE; metrics and averaging as in \cref{tab:main,tab:full}).
  Rows~(a)--(e) correspond to \cref{tab:velocity}.
  Image loss only: key transport trained without depth, matching, or velocity-curriculum supervision.
  Small numbers denote changes relative to each block's reference (green: better, red: worse, gray: within the 90\% bootstrap interval).
  $^\dagger$Keys transported using ground-truth content motion during training and testing.
  $^\ddagger$Ground-truth depth without key transport during training and testing.}
  \label{tab:ablations}
\end{table*}

\paragraph{Supervision.}
Trained with $\mathcal{L}_\text{img}$ alone (without depth losses, matching losses, or the velocity curriculum), key transport still improves \psnrd by 0.56\,dB over~(a), recovering 60\% of the full \method gain, though LPIPS and \lpipsd degrade slightly (\cref{tab:ablations}).

\paragraph{Velocities as token features.}
To check that the improvements come from placing keys at the query time and not just from exposing motion cues to the network, we feed the predicted velocities into the input tokens as additional features without transporting keys.
This variant is 0.72\,dB behind \method in \psnrd. Its +0.22\,dB gain over~(a) is on par with that of supervising the matcher without transporting keys (+0.23\,dB), and both variants raise static and full-image PSNR by roughly 0.2\,dB (\cref{tab:ablations}).
Token features are the same for every query, whereas key transport adapts each key's 3D position to the query timestamp. Motion cues therefore benefit dynamic rendering mainly through geometric key placement.

\paragraph{Disabling transport at test time.}
Setting velocities to zero at test time reduces \psnrd by 0.98\,dB (returning to the baseline level of~(a)) and lowers full-image PSNR by 0.22\,dB.
For the oracle model trained with ground-truth motion and depth, disabling transport lowers \psnrd by 2.69\,dB and full-image PSNR by 0.75\,dB (\cref{tab:ablations}).

\paragraph{Auxiliary depth losses.}
Without key transport, adding the dynamic weighting and the two regularizers of the depth loss changes RayRoPE by at most 0.10\,dB across all PSNR metrics (first two rows of \cref{tab:ablations}).

\begin{figure}[!htb]
  \centering
  \includegraphics[width=\linewidth]{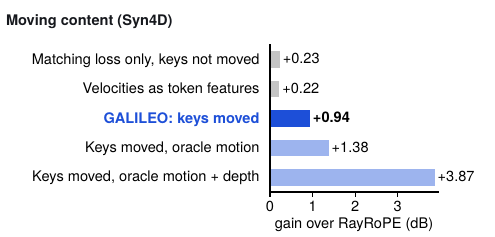}
  \caption{\textbf{Impact of key placement accuracy on Syn4D.} Gain in \psnrd over the RayRoPE baseline (19.88\,dB), averaged equally over novel views and novel times (\cref{tab:motion}).}
  \label{fig:headroom}
\end{figure}

\paragraph{Oracle motion and depth.}
\cref{tab:ablations} reports complete image metrics for the oracle variants in \cref{tab:motion}, and \cref{fig:headroom} summarizes their gains in \psnrd.
Replacing estimated velocities with ground-truth motion adds a further 0.44\,dB in \psnrd and 0.15\,dB in full-image PSNR over \method.
Ground-truth depth improves full-image PSNR much more than ground-truth motion does (+1.92\,dB vs.\ +0.49\,dB over~(a)), and combining the two adds another +2.3\,dB in \psnrd over ground-truth depth alone.

\paragraph{Input warping baselines.}
In the warping baselines of \cref{tab:motion,tab:ablations}, input images are warped to the query time (with their timestamps set to it) before entering RayRoPE instead of transporting keys.
\emph{GT warp} uses ground-truth depth, per-token ground-truth 3D displacements, and a ground-truth dynamic mask to shift only the moving pixels within each patch. Even so, it matches oracle \method only in \psnrd and is worse on \ssimd, \lpipsd, static and full-image PSNR, and all temporal metrics.
\emph{Predicted warp} uses the per-token depths and velocities predicted by a frozen 20k-step \method model, without a dynamic mask. Its full-image LPIPS is slightly lower, but it trails \method on every PSNR metric, \ssimd, \lpipsd, and JEDi.

\begin{table}[!htb]
  \centering
  \footnotesize
  \setlength{\tabcolsep}{3pt}
  \resizebox{\columnwidth}{!}{%
  \begin{tabular}{@{}l cc ccc@{}}
    \toprule
    & \multicolumn{2}{c}{image} & \multicolumn{3}{c}{video} \\
    \cmidrule(lr){2-3} \cmidrule(l){4-6}
    Model & PSNR$_\mathrm{dyn}\!\uparrow$ & PSNR$\uparrow$ & tOF$\downarrow$ & $\delta_\mathrm{avg}\!\uparrow$ & JEDi$\downarrow$ \\
    \midrule
    URoPE & 19.66 & 22.99 & 2.18 & .627 & 1.410 \\
    + \method, no detach & 20.10\,\up{+0.44} & 23.10\,\up{+0.10} & 1.80\,\up{$-$0.38} & .636\,\up{+.009} & 0.779\,\up{$-$0.630} \\
    + \method, detach & 20.14\,\up{+0.49} & 23.11\,\up{+0.12} & 1.82\,\up{$-$0.37} & .637\,\up{+.010} & 0.787\,\up{$-$0.622} \\
    \bottomrule
  \end{tabular}}
  \caption{\textbf{Effect of detaching matching features in URoPE + \method} on Syn4D (20k steps; metrics as in \cref{tab:main}).
  The detached variant corresponds to \cref{tab:main}; small numbers denote changes from URoPE, all statistically significant.}
  \label{tab:detach}
\end{table}

\begin{table}[!htb]
  \centering
  \footnotesize
  \setlength{\tabcolsep}{3pt}
  \begin{tabular}{@{}l cc@{}}
    \toprule
    Model & PSNR$_\mathrm{dyn}\!\uparrow$ & PSNR$\uparrow$ \\
    \midrule
    RayRoPE & 25.56 & 32.14 \\
    \method & 26.39\,\up{+0.83} & 32.26\,\up{+0.13} \\
    \bottomrule
  \end{tabular}
  \caption{\textbf{Scaling to a 12-layer backbone on Stereo4D.}
  Both models follow the training protocol of \cref{tab:main} with twice as many transformer layers and an FFN width of 3072.
  Small numbers indicate changes over RayRoPE (both outside the 90\% clip-bootstrap interval).}
  \label{tab:deeper}
\end{table}

\paragraph{Larger backbone.}
With a backbone twice as deep (12 layers, FFN width 3072) on Stereo4D, \method still improves over RayRoPE, by 0.83\,dB in \psnrd and 0.13\,dB in full-image PSNR (\cref{tab:deeper}).

\begin{table}[!htb]
  \centering
  \footnotesize
  \setlength{\tabcolsep}{3pt}
  \begin{tabular}{@{}l rrr@{}}
    \toprule
    12-layer model & RayRoPE & \method & $\Delta$ \\
    \midrule
    Parameters & 150.02M & 150.27M & +0.2\% \\
    Inference GFLOPs (1 target) & 7,449 & 7,496 & +0.6\% \\
    Inference GFLOPs (11 targets) & 81,832 & 81,894 & +0.08\% \\
    Inference peak memory (11 targets) & 4.0\,GB & 4.1\,GB & +0.1\,GB \\
    Training time (s/step) & 4.0 & 4.5 & +12\% \\
    Training step memory & 18.5\,GB & 19.1\,GB & +3.5\% \\
    \bottomrule
  \end{tabular}
  \caption{\textbf{Computational cost} of 12-layer RayRoPE and \method models (FFN width 3072) on eight $256 \times 256$ Syn4D input views in mixed precision.
  FLOPs count matrix multiplications, convolutions, and scaled dot-product attention (excluding elementwise RoPE and transport operations); each test video renders 11 target frames.
  Training is benchmarked at batch size 12 on a single NVIDIA B300 GPU; step memory denotes peak activation memory during a forward--backward pass above pre-step allocations (excluding parameters and optimizer states).}
  \label{tab:cost}
\end{table}

\paragraph{Computational overhead.}
\cref{tab:cost} compares the parameter, FLOP, and memory costs of the 12-layer RayRoPE and \method models.
Because the pixel-refinement CNN runs only once per clip, on the input views, the inference FLOP overhead drops from 0.6\% for a single target view to 0.08\% for an 11-frame video.
During training, \method increases step time by 12\% and peak activation memory by 3.5\%.

\begin{table}[!htb]
  \centering
  \footnotesize
  \setlength{\tabcolsep}{2.5pt}
  \begin{tabular}{ll cc ccc}
    \toprule
    & & \multicolumn{2}{c}{image} & \multicolumn{3}{c}{video} \\
    \cmidrule(lr){3-4} \cmidrule(l){5-7}
    Steps & Method & PSNR$_\mathrm{dyn}\!\uparrow$ & PSNR$\uparrow$ & tOF$\downarrow$ & $\delta_\mathrm{avg}\!\uparrow$ & JEDi$\downarrow$ \\
    \midrule
    \multirow{3}{*}{20k} & RayRoPE & 19.88 & 23.17 & 2.19 & .639 & 1.075 \\
     & \method & 20.82 & 23.51 & 1.65 & .668 & 0.380 \\
     & $\Delta$ & \,\up{+0.94} & \,\up{+0.34} & \,\up{$-$0.53} & \,\up{+.029} & \,\up{$-$0.695} \\
    \cmidrule{1-7}
    \multirow{3}{*}{80k} & RayRoPE & 20.51 & 23.75 & 1.88 & .658 & 0.419 \\
     & \method & 21.15 & 23.98 & 1.58 & .679 & 0.230 \\
     & $\Delta$ & \,\up{+0.64} & \,\up{+0.23} & \,\up{$-$0.30} & \,\up{+.021} & \,\up{$-$0.188} \\
    \bottomrule
  \end{tabular}
  \caption{\textbf{Longer training on Syn4D.}
  Comparison between the 20k-step models of \cref{tab:main} and 80k-step runs (with curriculum transitions at $s_0 = 20\text{k}$ and $s_1 = 60\text{k}$), evaluated on the metrics of \cref{tab:main}.
  $\Delta$ denotes changes over RayRoPE, all statistically significant.}
  \label{tab:longer}
\end{table}

\paragraph{Longer training.}
Training for $4\times$ longer (80k steps; \cref{tab:longer}) narrows the \psnrd margin from 0.94 to 0.64\,dB, yet \method continues to significantly outperform RayRoPE across every image and video metric.
At 80k steps, RayRoPE still trails 20k-step \method in \psnrd and all three temporal metrics, although it is better in full-image PSNR (23.75 vs.\ 23.51\,dB).

\section{3D Tracking with ARROW}
\label{sec:supp_tracking}

\begin{figure}[!htb]
  \centering
  \includegraphics[width=\linewidth]{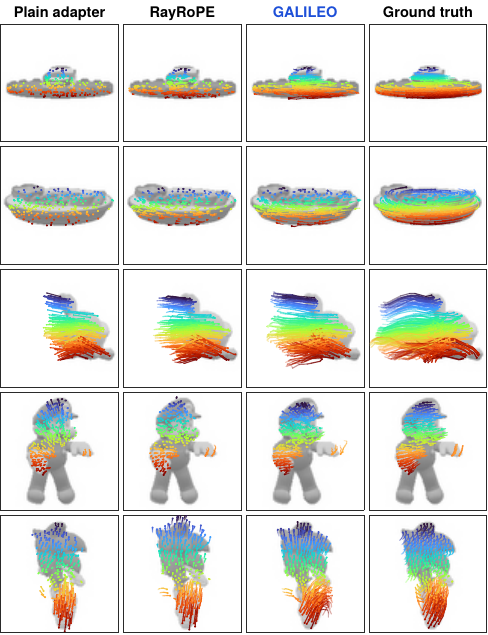}
  \caption{\textbf{More 3D tracks with ARROW} on five dynamic Objaverse test objects, in the format of \cref{fig:arrow_tracks}: inputs at 0 and 0.1\,s, tracks up to 0.3\,s, projected into the first camera and colored by initial height.}
  \label{fig:arrow_tracks_supp}
\end{figure}

This section details the 3D point-tracking adaptation experiments in \cref{sec:transfer} (\cref{tab:tracking,fig:arrow_tracks}).

\paragraph{Architecture and adapter.}
ARROW~\cite{fradlin2026arrow} encodes multi-view input images with a ViT alternating frame-wise and global attention, and predicts 3D point positions via a cross-attention query decoder.
Native ARROW takes no temporal input, answering novel-time queries using the nearest input frame of the target camera.
We freeze pretrained ARROW and insert lightweight attention adapters equipped with RayRoPE and temporal RoPE (rotating three rotary planes per key at $\omega = 2\pi$\,rad/s according to $t_b - t_a$); this adaptation is distinct from our use of off-the-shelf ARROW pseudo-tracks on Stereo4D (\cref{sec:supp_details}).
As a parameter-matched control, a plain adapter uses standard scaled dot-product attention without geometric or temporal encodings.
Queries sharing a target timestamp are grouped, and \method shifts each key's 3D points by $v_b\,(t_a - t_b)$. The velocities are estimated by the matcher of \cref{sec:velocity} on first-layer ARROW features: the lifted 3D displacement between the two input times (0.1\,s apart) is divided by 0.1\,s.
Both RayRoPE and \method adapters lift tokens via an auxiliary DPT depth head trained with stop-gradient to the frozen backbone.

\paragraph{Training.}
All adapters are trained identically for 5k steps with batch size $4$ over 40k training samples drawn from the 3,607 dynamic Objaverse training objects (disjoint from the 197 test objects), using ARROW's native tracking losses.
\method additionally applies $\mathcal{L}_\text{match}$ (\cref{sec:training}) and anneals from ground-truth to predicted velocities between $s_0 = 1.25\text{k}$ and $s_1 = 3.75\text{k}$ steps.
Each training sample provides two cameras at two timestamps 0.1\,s apart and supervises eight groups of 1,024 query points (half sampled on dynamic regions) spanning timestamps from 0 to 0.33\,s.

\paragraph{Evaluation.}
For each test object, 512 query points are sampled on the object surface across the four input views (half on moving parts) and tracked to one interpolated timestamp (0.03--0.06\,s) and five extrapolated timestamps (0.14--0.33\,s), evaluating points displaced by more than 1.5\,px (28,211 interpolated and 281,025 extrapolated queries).
At the two input timestamps, all query points are evaluated (201,728 queries), half of them at their own source timestamp, which isolates depth accuracy.
Predicted reconstructions are aligned per object in focal length and scale using the input-time predictions.
Both 3D end-point error (EPE) and average percent within threshold (APD, averaged over $\{1, 2, 4, 8, 16\}\%$) are normalized by the longest side of the object's bounding box.
\cref{fig:arrow_tracks} visualizes unaligned trajectories of 220 moving points on two test objects, projected into the first camera and color-coded by initial height.
\cref{fig:arrow_tracks_supp} shows five more test objects in the same format.

\section{Video Renderer for Code Worlds}
\label{sec:supp_agentgarten}
\begin{figure*}[t]
  \centering
  \includegraphics[width=\textwidth]{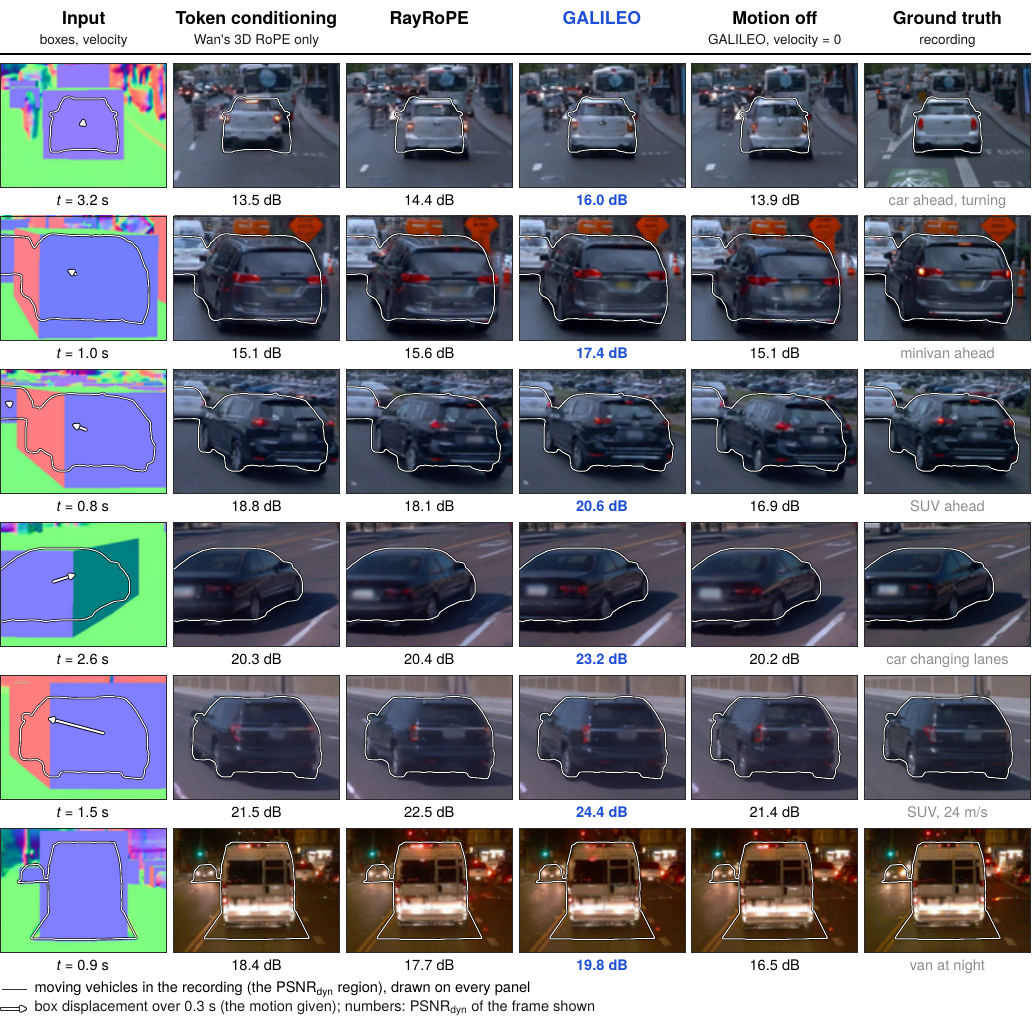}
  \caption{\textbf{Video renderer for code worlds: additional qualitative results} on held-out Waymo sequences conditioned on coarse 3D vehicle boxes.
  Token conditioning provides scene geometry only through conditioning tokens (\cref{tab:agentgarten}), while \emph{Motion off} evaluates the trained \method model with all velocities set to zero at test time.}
  \label{fig:driving_gallery}
\end{figure*}

This section details the driving video generation setup introduced in \cref{sec:transfer} (\cref{tab:agentgarten,fig:driving}), with additional qualitative comparisons presented in \cref{fig:driving_gallery}.

\paragraph{Architecture and conditioning.}
The released AgentGarten renderer~\cite{mirros2026agentgarten} builds on Cosmos3-Nano; we adopt its geometry-conditioning pathway and training pipeline but build on Wan2.2-TI2V-5B~\cite{wan2025wan} instead.
The generator receives a clean initial RGB frame, a structured text prompt specifying location, time of day, and weather, per-frame metric depth and surface normals, camera intrinsics and poses, and 3D vehicle bounding-box trajectories.
Metric depth is estimated by DA3-Metric-Large~\cite{lin2025da3} and aligned to LiDAR measurements, with surface normals derived from the resulting depth maps.
Depth maps are color-mapped, spatially downsampled by $4\times$, and encoded by the video VAE into 28 geometry tokens per latent frame.
Conditioned on the initial frame, the model synthesizes the subsequent 40 frames (4\,s at 10\,fps).

\paragraph{Coarse shape proxies.}
To emulate the coarse geometric primitives supplied by procedural or code-driven simulators, each vehicle's geometry during training is randomly represented by the raw DA3 surface, a low-pass-filtered surface, or its 3D bounding cuboid with probabilities $0.4$, $0.2$, and $0.4$, respectively.
At evaluation (\cref{tab:agentgarten}), all vehicles are represented only by 3D bounding cuboids.

\paragraph{Dataset and training.}
We use front-camera sequences from the Waymo Open Dataset~v2~\cite{sun2020waymo}, undistorted, cropped to $1920 \times 1108$, and resized to $832 \times 480$ at 10\,fps.
Models are fine-tuned for 3,000 steps at batch size 16 on 3,192 41-frame clips from 798 training segments, and evaluated on 223 clips from 126 validation segments containing 487 moving vehicle tracks.

\begin{figure*}[t]
  \centering
  \includegraphics[width=\textwidth]{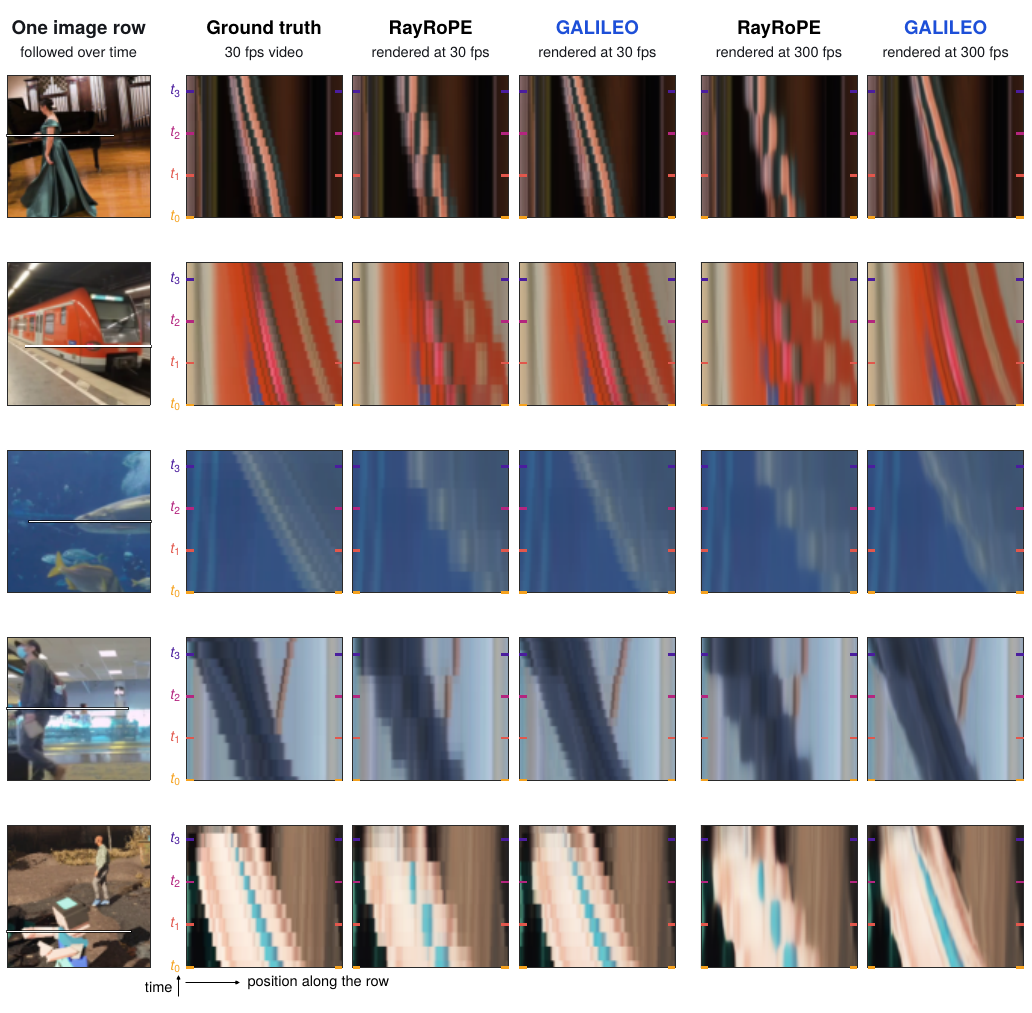}
  \caption{\textbf{Space--time slices} on five clips (following \cref{fig:xt}; the first row is the clip of \cref{fig:xt}). Right: RayRoPE and \method rendered at 300\,fps, $10\times$ the frame rate of the videos.}
  \label{fig:xt_two}
\end{figure*}

\paragraph{Compared variants.}
\emph{Token conditioning} (as in AgentGarten) supplies scene geometry only through conditioning tokens while retaining Wan's native 3D RoPE.
\emph{RayRoPE} additionally encodes near, middle, and far depth samples along each token's rays (rotating values as well as queries and keys).
\emph{\method} transports the three depth samples of each key token from key time $t_k$ to query time $t_q$ by $v\,(t_q - t_k)$ while holding the key camera center fixed, where $v$ is the world-space velocity of the ground-truth vehicle box dominating the token's footprint (computed via central differences over $\pm 1$ frame around $t_k$).
\emph{Motion off} evaluates the trained \method weights with all velocities set to zero at test time, reverting key positions to those of RayRoPE.

\begin{figure*}[t]
  \centering
  \includegraphics[width=\textwidth]{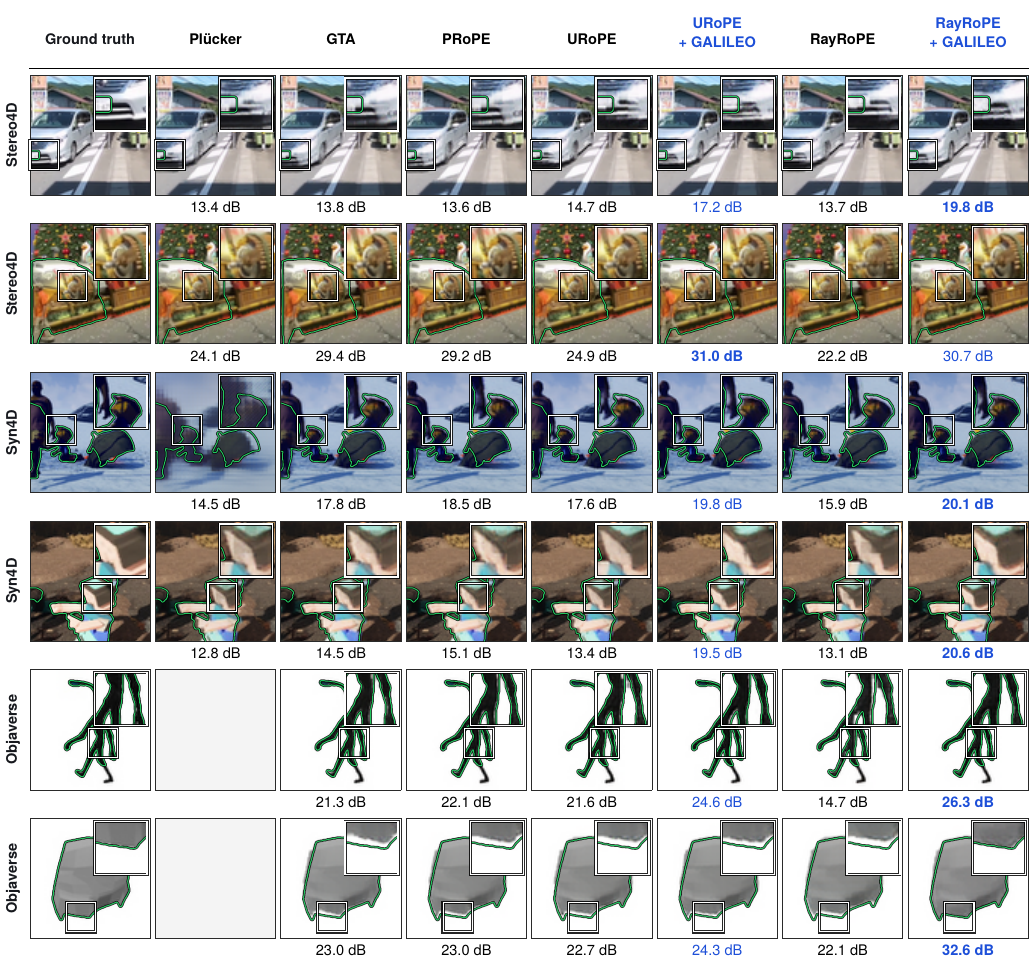}
  \caption{\textbf{Qualitative comparison across all encodings} on held-out frames from Stereo4D, Syn4D, and dynamic Objaverse at novel times (bottom row: novel view). Green contours outline ground-truth moving objects (top row: vehicle license plate), and values beneath each panel report frame-level \psnrd. Insets magnify the white-boxed regions; Pl\"ucker failed to converge on Objaverse.}
  \label{fig:qualitative}
\end{figure*}

\paragraph{Dynamic-region evaluation.}
A vehicle track is classified as moving if its median frame-to-frame box-center speed in world coordinates exceeds 1\,m/s, so that stationary objects are excluded even when the ego vehicle moves.
Its evaluation mask consists of the pixels whose back-projected recorded depth falls inside the vehicle's 3D bounding box (padded by 0.2\,m along length, width, and height, and truncated by 0.1\,m at the bottom to exclude road surfaces). This excludes occluders and isolates the visible vehicle surface.
Because the masks are derived from recorded LiDAR-aligned depth, they are identical for all conditioning variants and shape proxies, so the evaluation is on the true vehicle surface, even when models are conditioned on coarse cuboids.
Retaining tracks that span at least 1,500\,px in at least 8 of the 40 generated frames yields 487 moving vehicles across 223 clips.
Excluding the conditioning frame ($t = 0$), \psnrd is computed on generated frames 1--40 containing at least 100 dynamic pixels, averaged across frames within each clip and then across clips (and analogously within each 10-frame, 1\,s interval).

\paragraph{Silhouette IoU.}
To assess whether generated vehicles preserve accurate silhouettes and positions, we run a Cityscapes-pretrained~\cite{cordts2016cityscapes} Mask2Former~\cite{cheng2022mask2former} instance segmenter on every fourth generated frame.
For each moving vehicle, we select the predicted vehicle instance whose 2D bounding box overlaps the projected 3D vehicle box with an IoU of at least $0.3$, and compute mask IoU against ground-truth Waymo panoptic labels~\cite{mei2022waymopvps} (assigning 0 if no matching instance is detected).
Over all frames with ground-truth panoptic annotations (68 moving vehicles across 31 clips), the same detector reaches an IoU of 0.879 on the ground-truth recordings, which is an upper bound.

\section{Additional Qualitative Results}
\label{sec:supp_qualitative}

\cref{fig:xt_two} extends the space--time slice visualization of \cref{fig:xt} to five clips, and also renders both RayRoPE and \method at 300\,fps. \method reconstructs smooth, continuous space--time trajectories, whereas RayRoPE produces piecewise-constant staircases tied to input timestamps.
\cref{fig:qualitative} compares all positional encodings side by side on Stereo4D, Syn4D, and dynamic Objaverse. Integrating \method into both URoPE and RayRoPE consistently reduces ghosting and gives sharper moving details.

\section{Proof of \cref{prop:boost}}
\label{sec:supp_proof}

Consider a reference frame moving at constant velocity $-w$, in which any 3D point $x$ observed at time $t$ has coordinates $x + w t$.
Under this boost, a key point observed at $t_b$ becomes $x + w t_b$, while the query camera at $t_a$ has center $c_a + w t_a$ and unchanged orientation $R_a$.
When lifting correspondences with the shifted cameras, each 3D point at timestamp $t_n$ is displaced by $w t_n$; since the least-squares slope of $w t_n$ with respect to $t_n$ in \cref{eq:fit} is $w$, the estimated velocity becomes $v_x + w$.
The transported key point at $t_a$ is $(x + w t_b) + (v_x + w)(t_a - t_b) = \xh + w t_a$, whose query-camera coordinates remain unchanged: $R_a\T (\xh + w t_a - c_a - w t_a) = R_a\T (\xh - c_a)$.
In contrast, untransported coordinates shift by $-R_a\T w\,(t_a - t_b)$.
Because the intrinsics transformation $\tilde{K}_a$ in RayRoPE is linear, the same invariance holds for all ray coordinates.
With the true displacement in place of the least-squares estimate, $\xh = x(t_a)$, and the argument extends to any time-dependent rigid change of frame $g_t$: the query camera and the content at $t_a$ are both mapped by $g_{t_a}$, so $T_a(\xh)$ is unchanged.

\paragraph{Deviations from exact invariance.}
Two implementation choices leave a residual dependence on $w$.
First, the relative camera-center coordinates $R_a\T (c_b - c_a)$ (3 of the 12 coordinates in RayRoPE) shift by $R_a\T w\,(t_b - t_a)$ because observer positions are not transported.
Second, the first transformer layer operates with the untransported base encoding before velocity estimation (\cref{sec:supp_velocity}), so its attention weights and the resulting features fed into the matcher and depth heads vary slightly with $w$.
Removing both would require transporting each key's camera center together with its ray endpoints and estimating correspondences before the first layer; we keep the camera center fixed because it is the observer's position, not the content's (\cref{sec:transport}).
We quantify the residual sensitivity of our implementation in \cref{sec:supp_boost}.

\end{document}